\documentclass{article}

\usepackage[preprint]{neurips_2025}

\usepackage[utf8]{inputenc} 
\usepackage[T1]{fontenc}    
\usepackage{hyperref}       
\usepackage{url}            
\usepackage{booktabs}       
\usepackage{amsfonts}       
\usepackage{nicefrac}       
\usepackage{microtype}      
\usepackage{xcolor}         

\usepackage{amsmath, amssymb, bm}
\usepackage{graphicx}
\usepackage{makecell}
\usepackage{algorithm}
\usepackage{algpseudocode}
\usepackage{needspace}

\newcommand{\denoisernew}{g_{\theta}^{B}({\bf x}_t, t, {\bf y}_e)}

\usepackage[normalem]{ulem}
\useunder{\uline}{\ul}{}

\usepackage{todonotes}

\title{Binary Diffusion Probabilistic Model}

\author{%
  Vitaliy~Kinakh\thanks{Use footnote for providing further information
    about author (webpage, alternative address)---\emph{not} for acknowledging
    funding agencies.} \\
  Department of Computer Science\\
  University of Geneva\\
  Geneva, Switzerland \\
  \texttt{vitaliy.kinakh@unige.ch} \\
  \And
  Slava~Voloshynovskiy \\
  Department of Computer Science \\
  University of Geneva \\
  Geneva, Switzerland \\
  \texttt{svolos@unige.ch}
}

\begin{document}

\maketitle

\begin{abstract}

We propose the Binary Diffusion Probabilistic Model (BDPM), a generative framework specifically designed for data representations in binary form. Conventional denoising diffusion probabilistic models (DDPMs) assume continuous inputs, use mean squared error objectives and Gaussian perturbations, i.e., assumptions that are not suited to discrete and binary representations. BDPM instead encodes images into binary representations using multi bit-plane and learnable binary embeddings, perturbs them via XOR-based noise, and trains a model by optimizing a binary cross-entropy loss. These binary representations offer fine-grained noise control, accelerate convergence, and reduce inference cost. On image-to-image translation tasks, such as super-resolution, inpainting, and blind restoration, BDPM based on a small denoiser and multi bit-plane representation outperforms state-of-the-art methods on FFHQ, CelebA, and CelebA-HQ using a few sampling steps. In class-conditional generation on ImageNet-1k, BDPM based on learnable binary embeddings achieves competitive results among models with both low parameter counts and a few sampling steps.

\end{abstract}
\section{Introduction}

Generative models have become integral to advancements in modern machine learning, offering state-of-the-art (SOTA) solutions across various domains, including image synthesis, cross-modal tasks like text-to-image and text-to-video generation~\cite{rombach2022high, ho2022imagen}. Denoising diffusion probabilistic models (DDPMs)~\cite{sohl2015deep, ho2020denoising} are particularly prominent within this landscape, utilizing iterative noise-based transformations to generate high-quality samples. These models predominantly employ Gaussian-based diffusion, which, while effective for continuous data, is less suited to inherently discrete or binary data representations. Despite diffusion models’ initial development for binary and categorical data~\cite{sohl2015deep}, their adoption in these areas remains limited, leaving a gap for binary and discrete tasks in fields such as image processing and tabular data generation.

This paper introduces {\em Binary Diffusion Probabilistic Model (BDPM)}, a novel approach specifically tailored to binary representation of essentially non-binary discrete data, which extends diffusion processes to better capture the characteristics of binary structures. Unlike traditional DDPMs, that are applied to float representations of images, our BDPM model employs binary representations of the images. We investigate two types of the binary representations and test them in two applications of image-to-image translation and class conditional image generation. For image-to-image translation tasks, where per-pixel accuracy is important, we use bit-plane decomposition, while for class-conditional generation, where generic perceptual image quality is of importance, we deploy learnable binary representation. BDPM integrates a binary cross-entropy loss function, offering a binary similarity metric that enhances training stability and model convergence.

Our contributions are as follows: (i) {\em Novel Diffusion Generative Model}: We propose BDPM, a diffusion-based generative model designed for binary data representations, optimized for the requirements of binary structures. (ii) {\em State-of-the-Art Performance}: BDPM demonstrates superior performance across multiple image-to-image translation tasks, including super-resolution, inpainting, and blind image restoration, achieving competitive or improved results over existing SOTA approaches, including DDPM-based methods. BDPM demonstrates competitive class-conditional image generative performance on ImageNet-1k. (iii) {\em Small Size Model}. Our image-to-image translation models with only 35.8M parameters, outperforms larger models, that are often based on large text-to-image models or pretrained on large-scale datasets, in terms of speed and performance. Our conditional image generative model with 32.9M parameters demonstrates competitive performance to larger models. (iv) {\em Enhanced Inference Efficiency}: Our model attains high-quality results with a reduced number of sampling steps and small number of parameters.
By shifting from Gaussian to binary formulations in diffusion models, BDPM establishes a promising foundation for generative tasks where binary data representations are essential or beneficial from the computation and interpretation perspectives.

\section{Related work}

\textbf{Traditional DDPMs.} DDPMs \cite{sohl2015deep, ho2020denoising} have become the go-to solutions in generative modeling in the last years. These models define a forward diffusion process that progressively adds scaled Gaussian noise ${\boldsymbol \epsilon} \sim \mathcal{N}({\bf 0}, \mathbf{I})$ to data, transforming initially complex data distributions into a standard Gaussian distribution over multiple time steps. Specifically, the forward process is formulated as:
\begin{equation}
q(\mathbf{x}_t | \mathbf{x}_{t-1}) = \mathcal{N}\left(\mathbf{x}_t; \sqrt{1 - \beta_t} \, \mathbf{x}_{t-1}, \beta_t \, \mathbf{I}\right),    
\end{equation}
where $\mathbf{x}_t$ and $\mathbf{x}_{t-1}$ are the noisy data samples at time steps $t$ and $t-1$, respectively. $\beta_t$ is the variance schedule controlling the noise level at each time step $t$. Practically, $\mathbf{x}_t$ is computed as a mapping $\mathbf{x}_t = \sqrt{\bar{\alpha}_t} \, \mathbf{x}_0 + \sqrt{1 - \bar{\alpha}_t} \,\boldsymbol{ \epsilon}$, where $\mathbf{x}_0$ is the original data sample. $\bar{\alpha}_t = \prod_{s=1}^t (1 - \beta_s)$ is the cumulative product of $(1 - \beta_s)$ up to time $t$, representing the overall scaling factor due to noise addition. $\alpha_t = 1 - \beta_t$ is used for notational convenience.

The reverse denoising process aims to reconstruct the data by learning the reverse conditional distributions $p_\theta(\mathbf{x}_{t-1} | \mathbf{x}_t)$. This is achieved by training a neural network to predict the original noise added at each time step by minimizing the mean squared error (MSE) loss:

\begin{equation}
    \mathcal{L}_{\boldsymbol{ \epsilon}} = \mathbb{E}_{\mathbf{x}_0,\boldsymbol{ \epsilon} \sim \mathcal{N}({\bf 0}, \mathbf{I}), t} \left[ \left\|\boldsymbol{ \epsilon} - \hat{\boldsymbol{ \epsilon}} \right\|^2 \right],
\label{eq:ddpm_loss}
\end{equation}
where $\hat{\boldsymbol{\epsilon}} = {g}_{\theta}({\bf x}_t, t)$ is the noise predicted at timestep $t$ by the denoiser ${g}_{\theta}$, parameterized by $\theta$.

\textbf{Image representation}. In traditional DDPMs, a discrete image ${\bf I}_0 \in \mathcal{I}^{h \times w \times c}$ of size $h \times w \times c$, where $h$ and $w$ denote the height and width of the image, and $c$ represents the number of color channels, is represented as continuous-valued tensors to use the Gaussian diffusion process effectively. The continuous nature of the data and noise ensures that the loss function (\ref{eq:ddpm_loss}) provides meaningful gradients for learning. However, this approach assumes that the underlying data distribution is continuous, which is not a case for inherently discrete original data, such as images. When dealing with 8-bit images or discrete representations of images, representing them as continuous variables can be inefficient. The mismatch between the continuous and the discrete data distribution assumptions highlights the need for alternative diffusion models that can handle discrete data representations more effectively. 

\textbf{Binary DDPMs.} Binary Latent Diffusion (BLD)~\cite{wang2023binary} proposes a diffusion model operating in a binary latent space for image generation. Paper advocates an autoencoder that encodes images into binary Bernoulli-distributed latent vectors and a diffusion model is then trained on these binary latents to model their distribution. BLD focuses on modeling binary latent representations through a dedicated Bernoulli diffusion process, relying heavily on a specific autoencoder architecture and binary sampling mechanisms. While effective for generation tasks, its design is tightly coupled to its latent space, limiting flexibility across domains and data representations. In contrast, BDPM offers a more general framework capable of operating on arbitrary binary representations, including handcrafted bit-planes or learned embeddings. Thanks to XOR-based noise modeling, BDPM provides precise control over the corruption process and allows fine-grained manipulation at the bitplane level.

\textbf{Binary latent representations}. MAGViT-v2~\cite{yu2023language} proposes an embedder that maps images into concise binary tokens via lookup-free quantization scheme. When plugged into large language models, MAGViT-v2 outperforms existing SOTA models on image and video generation benchmarks.
MaskBit~\cite{weber2024maskbit} proposes a two-stage class-conditional image generator that uses binary bit tokens and trains a masked transformer to generate directly on these bit tokens.

Despite recent advances, most diffusion model research in image-to-image translation and image generation remains focused on continuous DDPMs, while investigations of binary image representations have largely relied on transformer‐based masked modeling or autoregressive generation. This dual emphasis has left a notable gap in the study of discrete diffusion processes tailored specifically to binary data. Adapting DDPM models to binary modalities is nontrivial and may fail to exploit the inherent advantages of binary embeddings. Binary DDPMs for image-to-image translation and image generation are still scarcely explored in the literature, representing a promising and timely direction for future research.
\section{Proposed method}

\subsection{Limitations of DDPMs}

DDPMs have achieved remarkable success in generating high-fidelity images~\cite{dhariwal2021diffusion} and have been extended to multiple tasks of image-to-image translation  such as super-resolution, inpainting, restoration and text-to-image generation, etc.~\cite{saharia2022palette}. However, their reliance on continuous data representations and Gaussian noise limits their applicability to inherently discrete or binary data, such as for example 8-bit RGB images or binary latent representations~\cite{yu2023language}. Below, we present key arguments for why a binary planes of {\em multi-bit plane representation} (MBPR), along with XOR-based noise addition, is a more natural choice for modeling digital images.

\textbf{Discrete nature of digital images.} Digital images in 8-bit RGB format are inherently discrete, with each color channel taking values from a finite set of 256 levels, corresponding to 8-bit binary representations. This natural discreteness motivates the adoption of a \emph{multi-bit plane image model}, in which each RGB channel is decomposed into 8 binary bit-planes. Each bit-plane thus forms a binary-valued image layer, with pixels taking values in $\{0, 1\}$. This representation provides a natural alignment with the underlying digital structure of the data and yields a more faithful modeling of image content compared to continuous-valued approximations.

In this work, we adopt the multi-bit plane image model as a foundational representation for various image-to-image translation tasks, including super-resolution, inpainting, and image restoration. In these settings, per-pixel accuracy is paramount, and our approach preserves the original image information in a lossless binary form, ensuring pixel-level consistency and structural fidelity.

For generative tasks, where the goal is to synthesize new content conditioned on latent variables, we employ a binarized latent representation, which aligns with the binary nature of our bit-plane modeling. Specifically, images are encoded into a compressed binary latent space, from which generation is performed. The resulting binary representation is then passed through a decoder, yielding high perceptual quality reconstructions. This strategy maintains consistency across encoding and generation while enabling efficient sampling and generation with reduced computational complexity. Importantly, our method achieves strong performance across perceptual and fidelity metrics, including FID, IS, Precision and Recall, while preserving structural details with low generation overhead.

\textbf{Incompatibility of Gaussian Noise with Discrete Representations.}  Gaussian noise, as applied in traditional DDPMs, assumes a continuous data space, suitable for real-valued data but not for binary or discrete values. When Gaussian noise is applied to binary data, intermediate values are generated, which must be quantized or binarized to maintain the binary format, potentially leading to artifacts, information loss and weak convergence. This shows a fundamental mismatch between Gaussian noise and the discrete structure of digital images.

\textbf{Suboptimality of MSE Loss in Discrete Space.}  DDPMs use MSE loss under a Gaussian-pixel assumption, training the denoiser to predict Gaussian noise rather than true discrete pixels~\cite{ho2020denoising}. This misalignment renders MSE suboptimal for discrete image data. We instead adopt binary embeddings (bit-planes or learnable embeddings) and a differentiable binary cross-entropy loss, which more accurately measures discrete reconstruction errors and consistently applies to both binary noise and image representations.

\subsection{Motivation for Binary Diffusion Probabilistic Models (BDPM)}

The proposed BDPM overcomes these limitations by adapting the diffusion process to binary data representation. BDPM applies noise to the binary representations (bit-planes or binary tokens) using XOR operations. By preserving the binary structure, BDPM enables loss functions that are better suited to discrete data, ultimately enhancing the model's performance on binary-specific tasks.

Therefore, while Gaussian-based DDPMs excel in image generation, their continuous nature limits their suitability for binary data. BDPM addresses this with binary representations and XOR-based noise, providing a solution better suited to digital image structure.

\subsection{Transform‐Domain Binary Representations}

Our approach relies on two binary embeddings (Figure~\ref{fig:fig1_bpr_latent}). For image‐to‐image translation: super‐resolution, inpainting and blind restoration we convert input tensors into non-learnable multiple bit-planes, ensuring pixel-level precision. For class-conditional generation, we employ a quantized autoencoder to produce a compact Learnable Binary Representation (LBR), enabling fast and efficient sampling without sacrificing perceptual fidelity.

\subsubsection{Non-learnable binary data representations: Multiple Bit-Plane Representation}

We aim at simple, tractable, and fully invertible data representations that are not learned but rather constructed through a deterministic transformation. Specifically, we employ the \emph{Multiple Bit-Plane Representation} (MBPR), which encodes image data using bit-plane decomposition. Given an image $\mathbf{I}_0$, we define a bijective, non-learnable transform $\mathcal{T}$ such that: $\mathbf{I}_0 = \sum_{k=0}^{n-1} \mathbf{x}_0(k) \cdot 2^k$, where each bit-plane $\mathbf{x}_0(k) \in \{0,1\}^{H \times W}$ captures the $k$-th binary coefficient across the entire image, with $k = 0, \ldots, n-1$, and $n$ is the bit-depth of the image.

Each bit-plane exhibits unique statistical properties: the most significant bits (MSBs) exhibit high inter-pixel correlation and structural fidelity, while the least significant bits (LSBs) are less correlated and more stochastic. This decomposition enables fine-grained control of noise injection during the diffusion process, allowing the model to prioritize denoising at the appropriate significance levels. The MBPR thus balances interoperability, computational complexity, and precision.

\subsubsection{Learnable binary data representations: Autoencoders with binarized latent space}

In contrast to MBPR, LBR are obtained via parametric encoders and decoders, typically trained end-to-end. One popular paradigm involves an encoder that transforms the image \(\mathbf{I}_0\) into a latent representation \(\mathbf{x}_0\), followed by a quantization stage. This is exemplified in architectures such as VQ-GAN~\cite{esser2021taming} or dVAE~\cite{rolfe2016discrete}, where \(\mathbf{x}_0\) represents quantized discrete tokens (e.g., integers via look-up table) or directly binary or represented in a binary form. In the current work, we proceed with MAGVIT-v2 \cite{yu2023language, luo2024open} with the explicit binary embeddings.

\begin{figure}[t]
    \centering
    \includegraphics[width=1.0\textwidth]{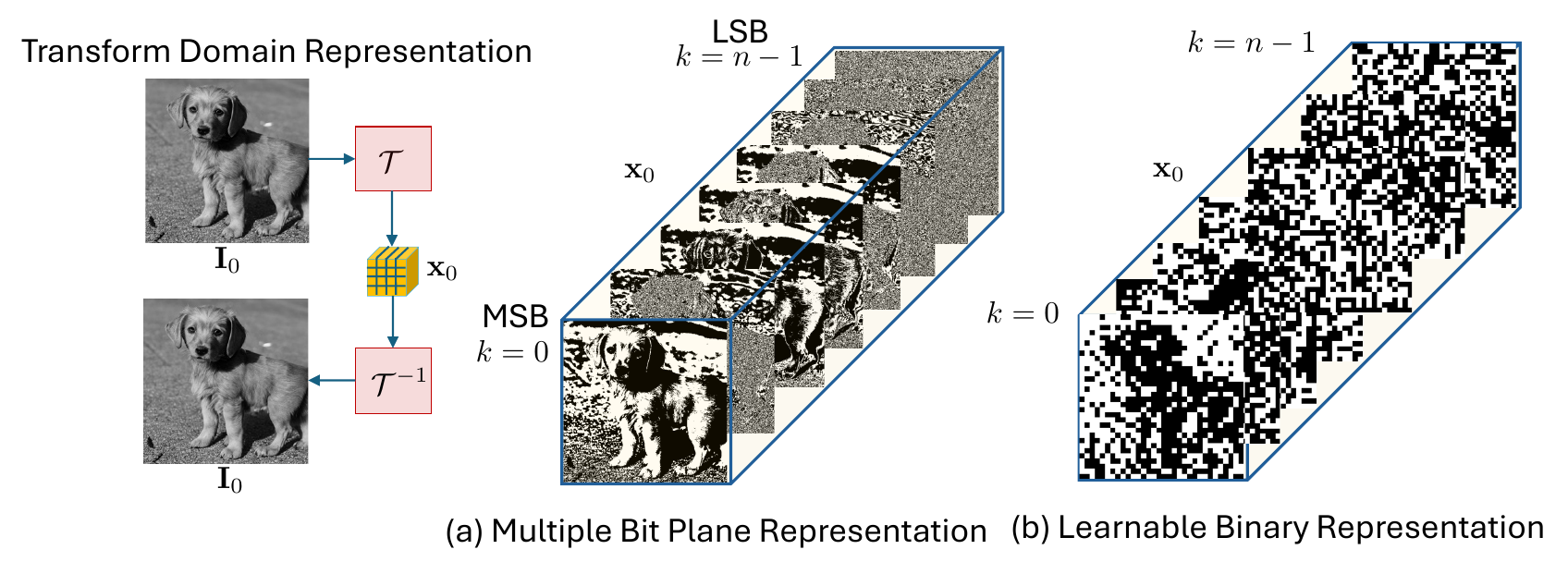}
    \caption{Transform domain binary data representations. \textbf{(a)}: \emph{MBPR} of an image \(\mathbf{I}_0\). The image is encoded using a bijective transform \(\mathcal{T}\) into a tensor \(\mathbf{x}_0\) of binary bit-planes, where MSB planes capture highly correlated structure and LSB planes contain decorrelated noise. \textbf{(b)}: \emph{LBR} obtained from an autoencoder architecture. The encoder outputs either quantized tokens which are subsequently binarized, or directly binarized latent vectors (e.g., via \(\{-1, +1\}\) projections).}
    \label{fig:fig1_bpr_latent}
\end{figure}

This class of lossy representation, illustrated in Figure~\ref{fig:fig1_bpr_latent}(b), maintains flexibility in capturing semantic features while enabling binary processing. The binary embeddings support interpretability, reduced storage, and efficient manipulation in downstream tasks, making it particularly attractive for integration into generative tasks. In this work, for tasks where perceptual per-pixel quality is important (super-resolution, inpainting, restoration) we use more expressive MBPR, for generative tasks with large data (ImageNet-1k generation) we use LBR.

\subsection{Binary Diffusion Probabilistic Model}

\begin{figure*}[t]
    \centering
    \includegraphics[width=\linewidth]{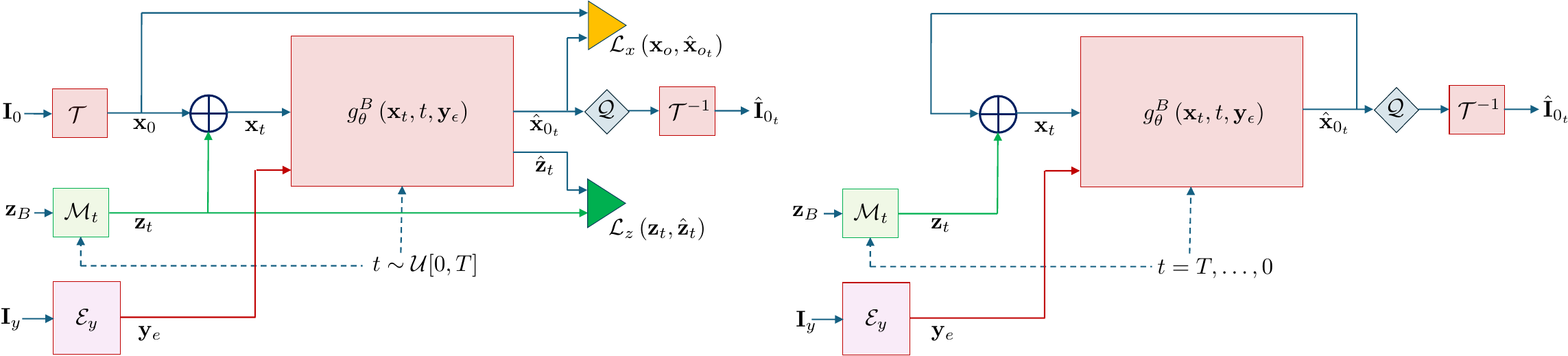}
    \caption{Binary Diffusion training (left) and sampling (right) schemes.}
    \label{fig:bd_train_sampling}
\end{figure*}

BDPM, shown in Figure~\ref{fig:bd_train_sampling}, is a novel approach for generative modeling that leverages the simplicity of binary data representations. This method involves adding noise through the XOR operation, which makes it particularly well-suited for handling binary data. Below, we describe the key aspects of the BDPM methodology in detail.

In BDPM noise is added to the data by flipping bits using the XOR operation as defined by the mapper $\mathcal{M}_t$ at each step $t$. The amount of noise added is quantified by the proportion of bits flipped. Let ${\bf x}_0 (k)\in \{0, 1\}^{h \times w}$ with $k=0,...,n-1$ for each image be the original channel of binary representation of dimension $h \times w$, and ${\bf z}_t(k) \in \{0, 1\}^{h \times w}$ be a random binary noise plane at time step $t$. The noisy binary channel ${\bf x}_t(k)$ at the output of $\mathcal{M}_t$ is obtained as: ${\bf x}_t(k) = {\bf x}_0(k) \oplus {\bf z}_t(k)$ , where $\oplus$ denotes the XOR operation. The noise level is defined as the fraction of bits flipped in ${\bf z}_t(k)$ in the mapper $\mathcal{M}_t$ at step $t$, with the number of bits flipped ranging within the probability range $[0, 0.5]$ as a function of the timestep and potentially as a function of $k$.

The denoising network $\denoisernew$ is trained to predict both the added noise tensor of bits ${\bf z}_t$ and the clean tensor of image binary representation ${\bf x}_0$ from the noisy tensor ${\bf x}_t$. We employ binary cross-entropy (BCE) loss for each bit plane to train the denoising network. The loss function is averaged over the batch of $M$ samples: 
\begin{equation*}
\label{eq:binary_diffusion_loss}
\begin{aligned}
\mathcal{L}(\theta) &= \frac{1}{B} \sum_{m=1}^M \left[ \mathcal{L}_{x}(\hat{{\bf x}}_0^{(m)}, {\bf x}_0^{(m)}) + \mathcal{L}_{z}(\hat{{\bf z}}_t^{(m)}, {\bf z}_t^{(m)}) \right], \\
\end{aligned}
\end{equation*}
where $\theta$ represents the parameters of the denoising network,  ${\bf x}_0^{(m)}$ and $\hat{{\bf x}}_0^{(m)}$ are the $m$-th samples of the true clean tensors and the predicted clean tensors, respectively. Similarly, ${\bf z}_t^{(m)}$ and $\hat{{\bf z}}_t^{(m)}$ are the $m$-th samples of the true added noise tensors and the predicted noise tensors, respectively. ${\bf y}_e = \mathcal{E}_y({\bf I}_y)$ denotes the encoded conditional image ${\bf I}_y$ that can represent the low-resolution down-sampled image, blurred image or image with removed parts that should be in-painted. The losses $\mathcal{L}_{x}$ and $\mathcal{L}_{z}$ denote BCE losses computed for each bit plane $k$ and the pixel coordinates $i\in\{1,\cdots,h\}$ and $j\in\{1,\cdots,w\}$ withing each bit plane. 

When MBPR representations are used, in order to balance bit-planes during training of the denoiser network, we apply linear bit-plane weighting, where the weight for MSB is set to 1, weight for LSB is set to 0.1 and for others weights are linearly interpolated between 1 and 0.1. When LBR representation are used, constant loss weight is used for all channels.

The output of the denoiser $\denoisernew$ is binarized via a mapper $\mathcal Q$ prior to apply the inverse transform ${\mathcal T}^{-1}$ as shown in Figure \ref{fig:bd_train_sampling}.

When sampling (Figure \ref{fig:bd_train_sampling} right), we start from a random binary tensor ${\bf x}_t$ at timestep $t = T$, along with the conditioning state ${\bf I}_y$, encoded into ${\bf y}_e$. For each selected timestep in the sequence $\{T, \ldots, 0\}$, denoising is applied to the tensor. The denoised tensor $\hat{{\bf x}}_0$ and the estimated binary noise $\hat{{\bf z}}_t$ are predicted by the denoising network. These predictions are then processed using a sigmoid function and binarized with a threshold in the mapper $\mathcal Q$. During sampling, we use the denoised tensor  $\hat{{\bf x}}_0$ directly. Then, random noise ${\bf z}_t$ is generated and added to $\hat{{\bf x}}_0$ via the XOR operation: ${\bf x}_t = \hat{{\bf x}}_0 \oplus {\bf z}_t$. The sampling algorithm is summarized in Algorithm \ref{alg:sampling}.

\begin{algorithm}
\caption{Sampling algorithm.}
\label{alg:sampling}
\begin{algorithmic}[1]
\State ${\bf x}_t \gets$ random binary tensor
\State ${\bf I}_{y} \gets$ condition/label
\State ${\bf y}_e \gets \mathcal{E}_y({\bf I}_{y})$ apply condition encoding 
\State $threshold \gets$ threshold value to binarize \Comment{Default 0.5}
\State $\denoisernew \gets$ pre-trained denoiser network

\For{$t \in \{T, \dots, 0\}$} \Comment{Selected timesteps}
 \State $\hat{{\bf x}}_0, \hat{{\bf z}}_t \gets \denoisernew$
    \State $\hat{{\bf x}}_0 \gets \sigma(\hat{{\bf x}}_0) > threshold$ \Comment{$\mathcal Q$: Apply sigmoid and compare to threshold}

    \State ${\bf z}_t \gets get\_binary\_noise(t)$ \Comment{Generate random noise}
    \State ${\bf x}_t \gets \hat{{\bf x}}_0 \oplus {\bf z}_t$ \Comment{Update ${\bf x}_t$ using XOR with ${\bf z}_t$}
\EndFor

\State \textbf{return} ${\mathcal T}({\bf x}_t)^{-1}$
\end{algorithmic}
\end{algorithm}

\section{Experimental results}

We evaluate the proposed method on: (a) image-to-image translation tasks such as 4× super-resolution task that scales images from $64 \times 64$ to $256 \times 256$ pixels using the FFHQ~\cite{karras2019style} and CelebA~\cite{liu2015faceattributes} datasets, medium size mask inpainting using FFHQ of the size $512 \times 512$ pixels, CelebA of the size $256 \times 256$ pixels, CelebA-HQ~\cite{karras2017progressive} of the size $512 \times 512$ and blind image restoration on CelebA of the size $256 \times 256$ and (b) conditional image generation on ImageNet-1k dataset~\cite{deng2009imagenet} of the size $256 \times 256$ pixels.

In image-to-image translation tasks,  we use the U-Net denoiser model with 35.8M parameters. 

For class-conditional image generation, we use binary latent representations, proposed in MAGVIT-v2~\cite{yu2023language}, in particular, we use the implementation from Open-MAGVIT2~\cite{luo2024open} with patch size of $16 \times 16$, and as a generator, we use the diffusion transformer proposed in DiT~\cite{peebles2023scalable}. We study 2 architectures: DiT-S with 32.9M parameters and DiT-B with 130M parameters. 

For all experiments, the sample inputs and outputs are provided in the Supplementary Materials.

\subsection{Metrics}

We evaluate the performance of image-to-image translation models using the Fréchet Inception Distance (FID)~\cite{heusel2017gans}, Learned Perceptual Image Patch Similarity (LPIPS)~\cite{zhang2018unreasonable}, Peak Signal-to-Noise Ratio (PSNR), Structural Similarity Index Measure (SSIM)~\cite{wang2004image}, and the number of sampling steps. For PSNR, SSIM and LPIPS we report mean value for the evaluation set. Perceptual Image Defect Similarity (P-IDS), Unweighted Image Defect Similarity (U-IDS)~\cite{zhao2021large} are also added to the inpainting evaluation. We evaluated the class-conditional image generation using FID~\cite{heusel2017gans}, Inception Score (IS)~\cite{salimans2016improved}, Precision, Recall~\cite{sajjadi2018assessing} and number of sampling steps. Number of sampling steps varies across evaluated tasks, since they are selected based on the best performance.

\subsection{Image-to-image translation tasks}

\subsubsection{Super-resolution}

The super-resolution process downscales the original images by selecting every 4-th pixel, then upsamples them back using bilinear interpolation. This upsampled image serves as the conditioning ${\bf I}_y$ (see Figure~\ref{fig:bd_train_sampling}). The upsampled conditioning image ${\bf I}_y$ is transformed into bit-planes via $\mathcal T$ and concatenated to the input ${\bf x}_t$. For super-resolution, we fix the number of sampling steps at 30. On FFHQ, BDPM outperforms GAN-based HiFaceGAN and diffusion methods DPS, DDRM, DiffPIR in LPIPS, PSNR and SSIM (Table~\ref{tab:ffhq_sr}). On CelebA, it surpasses GAN-based PULSE, diffusion models ILVR, DDNM, DifFace, ResShift, DiT-SR, and transformer-based VQFR, CodeFormer in FID, LPIPS and PSNR (Table~\ref{tab:celeba_sr}). With just 35.8 M parameters, BDPM achieves the best PSNR, LPIPS and SSIM, ranking among the best-performing yet smallest models.

\begin{table}[]

\centering

\caption{Comparison of super-resolution approaches on FFHQ dataset. The best metrics are shown in \textbf{bold} and second best {\ul underscored}. If the evaluation metric is not available in the paper, or in available public benchmark, it is marked as `-`.}
\label{tab:ffhq_sr}

\resizebox{0.9\columnwidth}{!}{

\begin{tabular}{lcccccc}
\hline
\textbf{Method} & \textbf{Params.} & \textbf{FID}  & \textbf{LPIPS} & \textbf{PSNR}  & \textbf{SSIM}  & \textbf{Steps} \\ \hline
HiFaceGAN~\cite{yang2020hifacegan} & 54M  & \textbf{5.36} & -              & 28.65          & 0.816          & 1    \\
DPS~\cite{chung2022diffusion}  &  554M  & 39.35         & 0.214          & 25.67          & 0.852          & 1000 \\
DDRM~\cite{kawar2022denoising} & 554M   & 62.15         & 0.294          & 25.36          & 0.835          & 1000 \\
DiffPIR~\cite{zhu2023denoising} & 554M   & 58.02         & 0.187          & 29.17          & -              & 20   \\
DiffPIR & 554M     & 47.8          & {\ul 0.174}    & {\ul 29.52}    & -              & 100  \\
BDPM (our) & 35.8M & {\ul 5.71}    & \textbf{0.151} & \textbf{30.05} & \textbf{0.864} & 30   \\ \hline
\end{tabular}
}
\end{table}

\begin{table}[]

\centering

\caption{Comparison of super-resolution approaches on CelebA dataset. The best metrics are shown in \textbf{bold} and second best {\ul underscored}. If the evaluation metric is not available in the paper, or in available public benchmark, it is marked as `-`.}
\label{tab:celeba_sr}

\resizebox{0.9\columnwidth}{!}{

\begin{tabular}{lcccccc}
\hline
\textbf{Method} & \textbf{Params.}  & \textbf{FID} & \textbf{LPIPS} & \textbf{PSNR}  & \textbf{SSIM}  & \textbf{Steps} \\ \hline
PULSE~\cite{menon2020pulse} & 26.2M & 40.33        & -              & 22.74          & 0.623          & 100  \\
DDRM~\cite{kawar2022denoising} & 554M  & 31.04        & -              & 31.04          & 0.941          & 100  \\
ILVR~\cite{choi2021ilvr} & 554M & 29.82        & -              & 31.59          & {\ul 0.943}    & 100  \\
VQFR~\cite{gu2022vqfr} & 76.3M & 25.24        & 0.411          & -              & -              & 1    \\
CodeFormer~\cite{zhou2022towards} & 74M  & 26.16        & 0.324          & -              & -              & 1    \\
DifFace~\cite{yue2024difface} & 176M & 23.21        & 0.338          & -              & -              & 100  \\
DDNM~\cite{wang2022zero} & 554M  & 22.27        & -              & {\ul 31.63}    & \textbf{0.945} & 100  \\
ResShift~\cite{yue2024resshift} & 118.6M  & {\ul 17.56}  & {\ul 0.309}    & -              & -              & 4    \\
DiT-SR~\cite{cheng2024effective} & 100.6M & 19.65        & 0.337          & -              & -              & 4    \\
BDPM (our) & 35.8M & \textbf{3.5} & \textbf{0.116} & \textbf{32.01} & 0.91 & 30   \\ \hline
\end{tabular}
}

\end{table}

\subsubsection{Inpainting}

The inpainting task reconstructs masked regions covering 10–30\% of each image. We convert the masked image~$\mathbf I_m$ into bit-planes, fill missing pixels with random $\{0,1\}$, and concatenate the mask~$\mathbf M$ and~$\mathbf I_m$ as $\mathbf I_y=[\mathbf M,\mathbf I_m]$. The denoiser runs for 100 sampling steps.

On FFHQ, BDPM outperforms LaMa, CoModGAN, TFill and SH-GAN in FID, PSNR, SSIM, P-IDS and U-IDS (Table~\ref{tab:ffhq_inpaint}), despite having the smallest model size.

On CelebA, BDPM surpasses RePaint, EdgeConnect, DeepFillV2, LaMa, DDRM, DDNM, ICT and MAT in FID, P-IDS and U-IDS (Table~\ref{tab:celeba_inpaint}), while remaining the second‐smallest model.

Evaluated on 10,000 CelebA-HQ images using the FFHQ-pretrained model, BDPM exceeds EdgeConnect, DeepFillV2, AOT GAN, MADF, LaMa, CoModGAN, ICT and MAT in FID and LPIPS (Table~\ref{tab:celeba_hq_inpaint}). 

\begin{table}[]

\centering

\caption{Comparison of inpainting approaches on FFHQ dataset. The best metrics are shown in \textbf{bold} and second best {\ul underscored}. If the evaluation metric is not available in the paper, or in available public benchmark, it is marked as `-`.}
\label{tab:ffhq_inpaint}

\resizebox{\columnwidth}{!}{
\begin{tabular}{lcccccccc}
\hline
\textbf{Method} & \textbf{Params.}  & \textbf{FID} & \textbf{LPIPS}  & \textbf{PSNR} & \textbf{SSIM}  & \textbf{P-IDS} & \textbf{U-IDS}   & \textbf{Steps} \\ \hline
LaMa~\cite{suvorov2022resolution} & 51M  & 19.6   & 0.287  & {\ul 18.99}     & {\ul 0.7178}  & -    & -      & 1    \\
CoModGAN~\cite{zhao2021large} & 109M & 3.7  & 0.247  & 18.46              & 0.6956           & {\ul 16.6} & {\ul 29.4}   & 1    \\
TFill~\cite{zheng2022bridging} & 70M   & 3.5  & \textbf{0.053}          & -             & -              & -              & -              & 1    \\
SH-GAN~\cite{xu2023image} & 79.8M & {\ul 3.4}          & 0.245          & 18.43         & 0.6936         & -              & -              & 1    \\
BDPM (our) & 35.8M & \textbf{1.3} & {\ul 0.059} & \textbf{28.7} & \textbf{0.961} & \textbf{17.43} & \textbf{33.07} & 100  \\ \hline
\end{tabular}
}

\end{table}

\begin{table}[]

\centering

\caption{Comparison of inpainting approaches CelebA dataset. The best metrics are shown in \textbf{bold} and second best {\ul underscored}. If the evaluation metric is not available in the paper, or in available public benchmark, it is marked as `-`.}
\label{tab:celeba_inpaint}
\resizebox{\columnwidth}{!}{

\begin{tabular}{lcccccccc}
\hline
\textbf{Method}  &  \textbf{Params.}  & \textbf{FID}  & \textbf{LPIPS} & \textbf{PSNR}  & \textbf{SSIM}  & \textbf{P-IDS} & \textbf{U-IDS} & \textbf{Steps} \\ \hline
RePaint~\cite{lugmayr2022repaint} &  554M  & 14.19         & -              & {\ul 35.2}     & {\ul 0.981}    & -              & -              & 2500             \\
DDRM~\cite{kawar2022denoising} & 554M  & 12.53         & -              & 34.79          & \textbf{0.982} & -              & -              & 100           \\
EdgeConnect~\cite{nazeri2019edgeconnect} & 27M  & 12.16         & -              & -              & -              & 0.84           & 2.31           & 1             \\
DeepFillV2~\cite{yu2019free} & 4.1M    & 13.23         & -              & -              & -              & 0.84           & 2.62           & 1             \\
ICT~\cite{wan2021high} & 120M  & 10.92         & -              & -              & -              & 0.9            & 5.23           & 1             \\
LaMa~\cite{suvorov2022resolution} & 51M & 8.75          & -              & -              & -              & 2.34           & 8.77           & 1             \\
MAT~\cite{li2022mat} & 62M & 5.16          & -              & -              & -              & {\ul 13.9}     & {\ul 25.13}    & 1             \\
DDNM~\cite{wang2022zero} & 554M & {\ul 4.54}          & -              & \textbf{35.64} & \textbf{0.982} & -              & -              & 100           \\
BDPM (our) & 35.8M & \textbf{1.96} & 0.08 & 28.3  & 0.928 & \textbf{15.04} & \textbf{27.01} & 100           \\ \hline
\end{tabular}
}

\end{table}

\begin{table}[]

\centering

\caption{Comparison of inpainting approaches on CelebA-HQ dataset. The best metrics are shown in \textbf{bold} and second best {\ul underscored}. If the evaluation metric is not available in the paper, or in available public benchmark, it is marked as `-`.}
\label{tab:celeba_hq_inpaint}

\resizebox{\columnwidth}{!}{

\begin{tabular}{lcccccccc}
\hline
\textbf{Method} & \textbf{Params.}  & \textbf{FID}  & \textbf{LPIPS} & \textbf{PSNR} & \textbf{SSIM} & \textbf{P-IDS} & \textbf{U-IDS} & \textbf{Steps} \\ \hline
EdgeConnect~\cite{nazeri2019edgeconnect} & 27M  & 10.58         & 0.101          & -             & -             & 4.14           & 12.45          & 1             \\
DeepFillv2~\cite{yu2019free} & 4.1M     & 10.11         & 0.117          & -             & -             & 3.11           & 9.52           & 1             \\
AOT GAN~\cite{zeng2022aggregated} & 15.2M        & 4.65          & 0.074          & -             & -             & 7.92           & 20.45          & 1             \\
MADF~\cite{zhu2021image} & 85M  & 3.39          & 0.068          & -             & -             & 12.06          & 24.61          & 1             \\
ICT~\cite{wan2021high} & 120M & 6.28          & 0.105          & -             & -             & 2.24           & 9.99           & 1             \\
LaMa~\cite{suvorov2022resolution} & 51M & 4.05          & 0.075          & -             & -             & 9.72           & 21.57          & 1             \\
CoModGAN~\cite{zhao2021large} & 109M  & 3.26          & 0.073          & -             & -             & {\ul 19.95}    & {\ul 31.41}    & 1             \\
MAT~\cite{li2022mat} & 62M & {\ul 2.86}    & {\ul 0.065}    & -             & -             & \textbf{21.15} & \textbf{32.56} & 1             \\
BDPM (our) & 36.8M & \textbf{1.17} & \textbf{0.06}  & 29.41 & 0.925 & 14.14           & 28.4           & 100           \\ \hline
\end{tabular}
}
\end{table}

\subsubsection{Blind image restoration}
Blind image restoration recovers high-quality images from unknown degradations. We pretrained BDPM on a synthetic FFHQ dataset, applying random perturbations. With 40 sampling steps, BDPM outperforms CodeFormer, DR2, BFRFormer, DiffBIR, GFP-GAN, BFRfusion, StableSR and DifFace in FID and SSIM (Table~\ref{tab:celeba_blind_image_restoration}). 

\begin{table}[]

\centering

\caption{Comparison of Blind Image Restoration on CelebA dataset. The best metrics are shown in \textbf{bold} and second best {\ul underscored}. If the evaluation metric is not available in the paper, or in available public benchmark, it is marked as `-`.}
\label{tab:celeba_blind_image_restoration}
\resizebox{0.87\columnwidth}{!}{

\begin{tabular}{lcccccc}
\hline
\textbf{Method} & \textbf{Params.} & \textbf{FID} & \textbf{LPIPS}   & \textbf{PSNR} & \textbf{SSIM}   & \textbf{Steps} \\ \hline
CodeFormer~\cite{zhou2022towards} & 74M  & 60.62       & 0.299  & 22.18         & 0.61            & 1             \\
DR2~\cite{wang2023dr2}  & 168M  & 58.94       & 0.3979       & 24.44         & 0.6784          & 250           \\
BFRFormer~\cite{ge2024bfrformer} & -   & 57.37       & \textbf{0.27}           & 22.83         & -               & 1             \\
DiffBIR~\cite{lin2023diffbir} & 17.2B   & 47.9        & 0.3786     & {\ul 25.6}    & 0.6809          & 50            \\
GFP-GAN~\cite{wang2021towards} & 56M  & 42.62       & 0.3646 & 25.08         & 0.6777          & 1             \\
BFRffusion~\cite{chen2024towards} & 1.23B  & 40.74       & 0.3621          & \textbf{26.2} & {\ul 0.6926}    & 50            \\
StableSR~\cite{wang2024exploiting} & 1.20B & 39.73       & 0.3637           & 24.84         & 0.6772          & 200           \\
DifFace~\cite{yue2024difface} & 176M  & {\ul 20.29} & 0.461   & 23.44         & 0.67            & 100           \\
BDPM (our) & 35.8M & \textbf{12.93} & {\ul 0.293} & 24.58 & \textbf{0.7415} & 40            \\ \hline
\end{tabular}

}
\end{table}

\subsection{Class-conditional image generation}
For class‐conditional generation, we condition on one‐hot labels and apply classifier‐free guidance~\cite{ho2022classifier}. We benchmark BDPM against comparable GANs (BigGAN, StyleGAN-XL), masked transformers (MaskGIT, MaskBit), autoregressive models (VQGAN, MAGVIT-v2, TiTok-S-128, LlamaGen, AiM, FAR-B), the continuous diffusion model LDM and the discrete diffusion model BLD. BDPM achieves competitive performance with low number of parameters and 7 sampling steps. As shown in Table \ref{tab:imagenet1k_generation}, BDPM-S achieves  FID of 8.93 and IS of 245.5, closely matching the discrete diffusion baseline BLD with only 7 sampling steps and 32.9M parameters. BDPM-B with 130M parameters, further reduces FID to 7.58 and elevates IS to 309.4, narrowing the gap to much larger autoregressive and diffusion models, while requiring only 7 sampling steps. FID and IS results can be explained by the fact that generated images are not very diverse. This can be partially explained by the fact, that, in order to make training faster, we have used cached precomputed features without augmentations. The detailed analysis of this phenomena and relationship between number of sampling steps and performance metrics are provided in the Supplementary Materials.

\begin{table}[]

\centering
\caption{Class-conditional generation of ImageNet-1k. If the evaluation metric is not available in the paper, or in available public benchmark, it is marked as `-`.}
\label{tab:imagenet1k_generation}

\resizebox{\columnwidth}{!}{

\begin{tabular}{llcccccc}
\hline
\textbf{Model type} & \textbf{Model}  & \textbf{Params.} & \textbf{FID}   & \textbf{IS} & \textbf{Precision} & \textbf{Recall} & \textbf{Steps} \\ \hline
GAN                & BigGAN~\cite{brock2018large}       & 112M   & 6.95  & 224.5  & 0.89      & 0.38   & 1     \\
                   & StyleGAN-XL~\cite{sauer2022stylegan}   & 166M   & 2.3   & 265.1  & 0.78      & 0.53   & 1     \\ \hline
Mask               & MaskGIT~\cite{chang2022maskgit}      & 227M   & 6.18  & 182.1  & 0.8       & 0.51   & 8     \\
                   & MaskBit~\cite{weber2024maskbit}      & 305M   & 1.52  & 328.7  & -         & -      & 256   \\
                   & MaskBit      & 305M   & 1.62  & 338.7  & -         & -      & 64    \\ \hline
AR                 & VQGAN~\cite{esser2021taming}        & 227M   & 18.65 & 80.4   & 0.78      & 0.26   & 256   \\
                   & MAGVIT-v2~\cite{yu2023language}    & 307M   & 1.91  & 324.3  & -         & -      & 64    \\
                   & Open-MAGVIT2~\cite{luo2024open} & 343M   & 3.08  & 258.3  & 0.85      & 0.51   & 256   \\
                   & TiTok-S-128~\cite{yu2024image}  & 287M   & 1.97  & -      & -         & -      & 64    \\
                   & LlamaGen-B~\cite{sun2024autoregressive}   & 111M   & 5.46  & 193.6  & 0.83      & 0.45   & 256   \\
                   & LlamaGen-L   & 343M   & 3.81  & 248.3  & 0.83      & 0.52   & 256   \\
                   & AiM-B~\cite{li2024scalable}        & 148M   & 3.52  & 250.1  & 0.83      & 0.52   & 256   \\
                   & AiM-L        & 350M   & 2.83  & 244.6  & 0.82      & 0.55   & 256   \\
                   & FAR-B~\cite{yu2025frequency}        & 208M   & 4.26  & 248.9  & 0.79      & 0.51   & 10    \\
                   & FAR-L        & 427M   & 3.45  & 282.2  & 0.8       & 0.54   & 10    \\ \hline
Diffusion          & LDM~\cite{rombach2022high}          & 400M   & 10.56 & 103.49 & 0.71      & 0.62   & 250   \\ \hline
Discrete Diffusion & BLD~\cite{wang2023binary}          & 172M   & 8.21  & 162.32 & 0.72      & 0.64   & 64    \\
                   & BDPM-S (our) & 32.9M  & 8.93  & 245.51  &  0.71 &  0.39 & 7     \\
                   & BDPM-B (our) & 130M   & 7.58  & 309.36  & 0.70      & 0.44   & 7     \\
                   \hline
\end{tabular}

}
\end{table}
\section{Limitations and Societal Impact}
BDPM requires task‐specific tuning of its noise schedule, loss weights and number of sampling steps, and has only been tested on 8-bit RGB images up to $512 \times 512$, leaving its performance on higher bit-depths or other modalities unverified. BDPM can improve image quality and generation efficiency but also risks deepfake misuse. Responsible use and countermeasures such as watermarking and fake detection tools are recommended.

\section{Conclusions}
We present BDPM, a novel framework that replaces continuous representations and Gaussian noise with multi bit-plane or learnable binary embeddings, XOR-based perturbations, and a binary cross-entropy loss. This discrete formulation better aligns with the intrinsic structure of binary data, yielding faster convergence and more precise denoising in the transform domain.

Empirically, BDPM achieves SOTA results on image-to-image translation—super-resolution, inpainting, and blind restoration on FFHQ, CelebA and CelebA-HQ—with just 35.8 M parameters and few sampling steps. In class-conditional ImageNet-1k generation, BDPM-S (32.9 M) attains FID 8.93 and IS 245.5 in 7 steps, while BDPM-B (130 M) further pushes to FID 7.58 and IS 309.4 in the same budget, rivaling much larger models. BDPM’s lightweight design and reduced sampling requirements make it especially well suited for deployment on resource-constrained hardware.


{
\small
    \bibliographystyle{ieeenat_fullname}
    \bibliography{main}
}


\appendix

\section{Datasets}

We perform experiments on four datasets: ImageNet-1k~\cite{deng2009imagenet}, CelebA~\cite{liu2015faceattributes}, FFHQ~\cite{karras2019style}, and CelebA-HQ~\cite{karras2017progressive}. ImageNet-1k comprises approximately 1.28 million natural images spanning 1,000 object categories, with an average resolution of $500 \times 500$ pixels. For training the class-conditional image generation model, we employ precomputed cached features to accelerate optimization; the training set is used for parameter updates, while the validation set monitors loss for convergence. During evaluation, we sample 50,000 images from the validation set to compute FID, IS, Precision and Recall. CelebA contains 202,599 celebrity images at approximately $178\times218$ pixel resolution, each annotated with 40 facial attributes; we use the training split for model fitting and the test split for evaluation. FFHQ comprises 70,000 high-quality face images at $1024\times1024$ resolution, offering diverse variations in age, ethnicity, and background; we allocate 60,000 images for training and 10,000 for evaluation. CelebA-HQ, a high-quality variant of CelebA, comprises 30,000 images at $1024\times1024$ resolution. For the super-resolution task, we utilize CelebA and FFHQ; the inpainting task employs CelebA, FFHQ, and CelebA-HQ. For blind image restoration, our model is pretrained on FFHQ and evaluated on CelebA.

\section{Implementation details}

{\bf Training setup}: We employ the PyTorch~\cite{Ansel_PyTorch_2_Faster_2024} and Accelerate~\cite{accelerate} frameworks for both training and inference. To accelerate these processes, we integrate FlashAttention~\cite{dao2022flashattention} and bfloat16 precision. During training, the denoiser predicts both the clean image’s binary representation and the added binary noise, by predicting the output tensor with twice number of channels of input, where first half corresponds to image representation and second half corresponds to noise. At each timestep \(t\in[0,T]\), the amount of noise added to each channel of binary representation remains constant. We define the noise level as the fraction of bits flipped in \(\boldsymbol{z}_t(k)\) by the mapper \(\mathcal{M}_t\) at step \(t\), with the flip probability ranging from 0 to 0.5 as a function of \(t\). To regulate this noise level, we adopt a quadratic noise schedule for the diffusion process. The schedule \(\beta_t\) at time step \(t\) is defined as
\[
\beta_t = \Bigl(\sqrt{\beta_{\mathrm{start}}} + \frac{t}{T}\bigl(\sqrt{\beta_{\mathrm{end}}}-\sqrt{\beta_{\mathrm{start}}}\bigr)\Bigr)^2,
\]
where \(T\) denotes the total number of timesteps (default: 1000), \(\beta_{\mathrm{start}}\) is the minimum noise value (\(10^{-5}\) by default), and \(\beta_{\mathrm{end}}\) is the maximum noise value (\(0.5\) by default). Here, \(\beta_t\) controls the bit-flip probability.

We apply exponential moving average (EMA) distillation to the denoiser model, using the EMA weights for evaluation. The final model is used for all evaluation.

The training parameters for each denoiser model are summarized in Table~\ref{tab:training_hyperparameters}.

{\bf Image-to-image translation denoiser architecture}: For image-to-image translation tasks, we employ the U-Net~\cite{ronneberger2015unet} denoiser network, which offers a balance between quality and speed. This architecture is inspired by the denoiser proposed in~\cite{dhariwal2021diffusion}. The U-Net comprises four convolutional downsampling blocks: the deepest block incorporates self-attention~\cite{vaswani2017attention}, while linear attention~\cite{katharopoulos2020transformers} is employed in the remaining blocks. The denoiser comprises 35.8M parameters. Sinusoidal timestep conditioning~\cite{ho2020denoising} is integrated as additive biases in every block. Image conditioning is incorporated by appending the input image as additional channels. During training, we adopt a linearly interpolated weighting scheme for the binary cross-entropy loss to account for the varying importance of different bit-planes of image: the most significant bit-plane is assigned a weight of 1, the least significant bit-plane a weight of 0.1, and intermediate bit-planes receive linearly interpolated weights. This approach ensures that higher bit-planes, which contribute more to the image’s overall structure and quality, are prioritized. The weights for noise prediction remain constant at 1 for each bit-plane.

{\bf Class-conditional image generation denoiser architecture}: We adopt the DiT~\cite{peebles2023scalable} architecture for conditional image generation. We evaluate two variants: DiT-S with 32 M parameters and DiT-B with 130 M parameters. Class conditionings are processed through an embedding layer and added as additive biases to every block. As the binary representation, we employ embeddings produced by the MAGVIT-v2 encoder with a patch size of $16\times16$~\cite{luo2024open}. During training, we follow initialization and update of parameter of MAGVIT-v2, applying weight decay only to weight parameters, excluding bias and positional embedding layers. The model is pretrained for 1,000,000 steps using a cosine learning rate schedule with 50,000 warmup steps. During sampling, we employ classifier-free guidance. 

\begin{table}[h]
\centering

\caption{Training Hyperparameters}
\label{tab:training_hyperparameters}

\begin{tabular}{@{}lccc@{}}
\toprule
\textbf{Hyperparameter} & \textbf{U-Net}  & \textbf{DiT-S} & \textbf{DiT-S} \\
\midrule
Optimizer & AdamW & AdamW & AdamW  \\
Learning rate & $1 \times 10^{-4}$ &  $4 \times 10^{-4}$ & $4 \times 10^{-4}$ \\
Weight decay & $1 \times 10^{-6}$ & $1 \times 10^{-2}$ & $1 \times 10^{-2}$  \\
Number of training steps & 500,000 & 1,000,000 & 1,000,000 \\
EMA update frequency & 10 steps & 10 steps & 10 steps \\
EMA decay & 0.995 & 0.995 & 0.995 \\
Noise schedule & Quadratic & Quadratic & Quadratic \\
Number of diffusion steps & 1,000 &  1,000 & 1,000 \\
Image bit-plane weights & Linear & Constant  & Constant \\
Mask bit-plane weights & Constant  & Constant  & Constant \\
Total batch size & \makecell[c]{64 for $256 \times 256$ \\ 32 for $512 \times 512$} &  2048 & 2048 \\
Hardware & \makecell[c]{1 L40S for $256 \times 256$ \\ 4 L40S for $512 \times 512$} & 8 RTX4090 & 8 L40S \\
Training time & \makecell[c]{12 hours for $256 \times 256$ \\ 44 hours for $512 \times 512$} & 45 hours & 148 hours \\
Classifier-free guidance scale & - & 11.25 & 8.75 \\ \hline

\bottomrule
\end{tabular}
\end{table}

\textbf{Super-Resolution Implementation}: Conditioning is performed by concatenating the bit-planes of the bilinearly upsampled low-resolution image as additional channels. During training, random cropping (80\%–100\% of the image height and width) and horizontal flipping are applied for data augmentation. For evaluation, we employ the Fréchet Inception Distance (FID)~\cite{heusel2017gans}, Learned Perceptual Image Patch Similarity (LPIPS)~\cite{zhang2018unreasonable}, Peak Signal-to-Noise Ratio (PSNR), Structural Similarity Index Measure (SSIM)~\cite{wang2004image} metrics.

\textbf{Inpainting Implementation}: Conditioning is performed by concatenating the bit-planes of masked images, where masked pixels are replaced with random noise, and by appending the mask as an additional channel. During training, random cropping (80\%–100\% of the image height and width) and horizontal flipping are applied for data augmentation. For evaluation, we employ the PSNR, SSIM, LPIPS, FID, Perceptual Image Defect Similarity (P-IDS), Unweighted Image Defect Similarity (U-IDS)~\cite{zhao2021large} metrics.

\textbf{Blind Image Restoration Implementation}: Conditioning is performed by concatenating the bit-planes of the perturbed image as additional channels. We pretrain our BDPM model on a synthetic blind image restoration dataset constructed from FFHQ images at \(256\times256\) resolution. Degradations are simulated by randomly applying a combination of perturbations, as summarized in Table~\ref{tab:perturbations}.

\textbf{Class-conditional Image Generation Implementation}: To accelerate training, we precompute the MAGVIT-v2 binary embeddings on the ImageNet-1k dataset without augmentations. For evaluation, we employ the FID, IS, Precision, and Recall metrics.

\begin{table}[h]
\centering

\caption{Perturbations, Parameters, and Probabilities Used in the Degradation Process.}
\label{tab:perturbations}


\begin{tabular}{l l c}
\toprule
\textbf{Perturbation} & \textbf{Parameters} & \textbf{Probability} \\
\midrule
Gaussian Blur & 
\makecell[l]{Kernel size: $21 \times 21$ \\ 
Kernel: isotropic or anisotropic \\ 
$\sigma_x, \sigma_y \in [0.1, 7]$ \\ 
Rotation angle $\in [-\pi, \pi]$} & 
Always \\
\addlinespace
Downsampling & 
\makecell[l]{Scale $[1, 4]$} & 
Always \\
\addlinespace
Additive Gaussian Noise & 
\makecell[l]{Standard deviation $[0, 15]/255$} & 
Always \\
\addlinespace
JPEG Compression & 
\makecell[l]{Quality factor $[50, 100]$} & 
Always \\
\addlinespace
Color Shift & 
\makecell[l]{Shift per color $[-20/255, 20/255]$} & 
30\% \\
\addlinespace
Color Jitter & 
\makecell[l]{Brightness $[0.5, 1.5]$ \\ 
Contrast $[0.5, 1.5]$ \\ 
Saturation $[0, 1.5]$ \\ 
Hue $[-0.1, 0.1]$} & 
30\% \\
\addlinespace
Grayscale Conversion & 
--- & 
1\% \\
\bottomrule
\end{tabular}
\end{table}

\section{Comparison with Posterior Sampling Methods}
BDPM applied to image-to-image translation tasks (super-resolution, inpainting, and blind image restoration) is compared to posterior sampling models such as DPS~\cite{chung2022diffusion}, DDRM~\cite{kawar2022denoising}, DiffPIR~\cite{zhu2023denoising}, DDNM~\cite{wang2022zero}, DiffBIR~\cite{lin2023diffbir}, BFRFusion~\cite{chen2024towards}, and StableSR~\cite{wang2024exploiting}. These methods leverage large pretrained diffusion models~\cite{dhariwal2021diffusion,rombach2022high} and adapt them for specific tasks via posterior sampling. Although they do not require additional training, they often necessitate more sampling steps and rely on larger models for inference, making them less suitable for large-scale deployment. In contrast, the proposed BDPM method requires training but is compact, with 35.8M parameters, and enables fast sampling. It outperforms posterior sampling methods, rendering it more suitable for large-scale applications.

\section{Effect of number of sampling steps}

\subsection{Image-to-image translation}

\begin{figure*}
    \centering

    \begin{minipage}{0.37\textwidth}
        \centering
        \includegraphics[width=\linewidth]{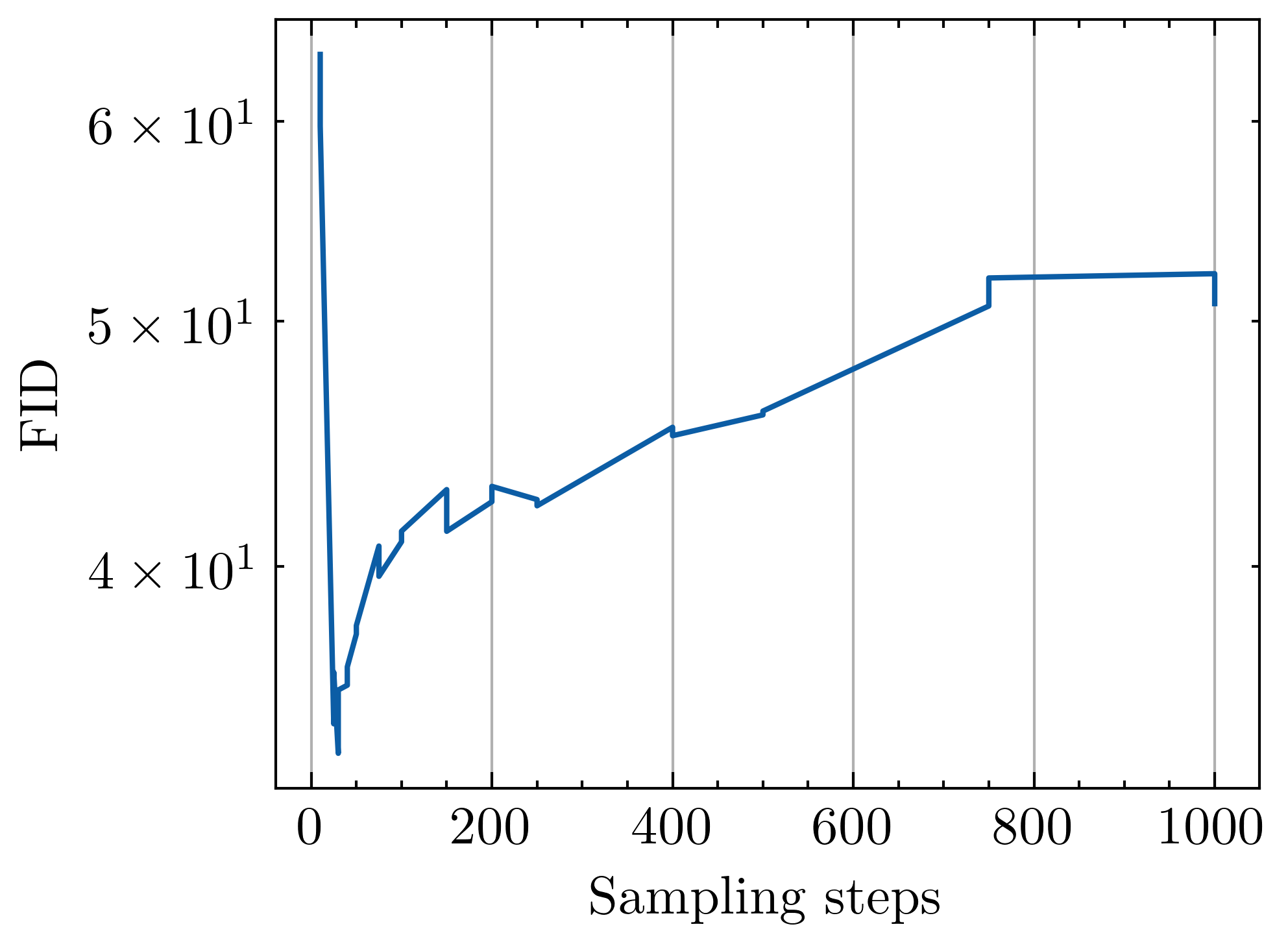}
        \par (a) FID 
        \label{fig:ffhq_sr_fid}
    \end{minipage}
    \hspace{1cm}
    \begin{minipage}{0.37\textwidth}
        \centering
        \includegraphics[width=\linewidth]{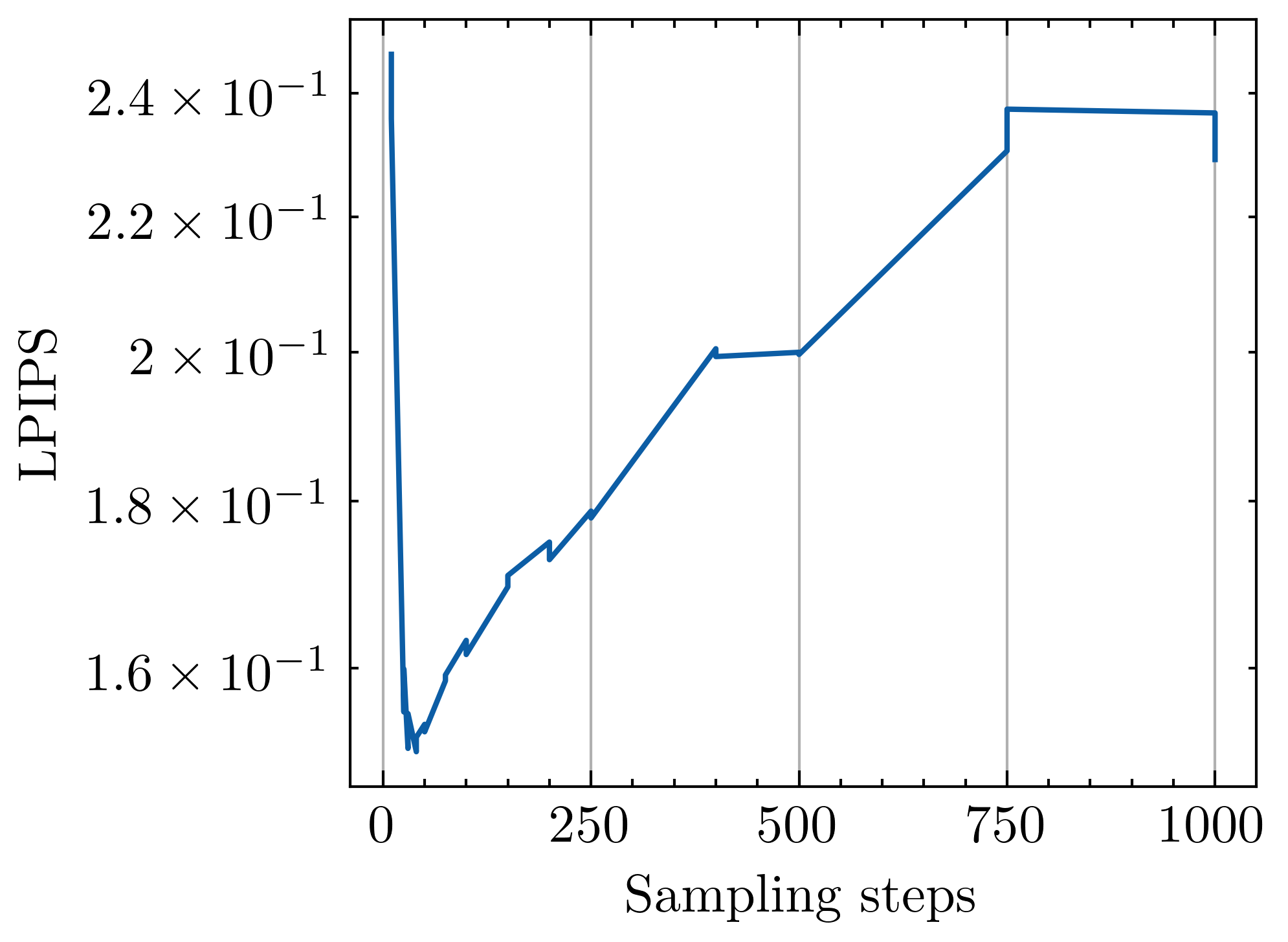}
        \par (b) LPIPS
        \label{fig:ffhq_sr_lpips}
    \end{minipage}

    \begin{minipage}{0.37\textwidth}
        \centering
        \includegraphics[width=\linewidth]{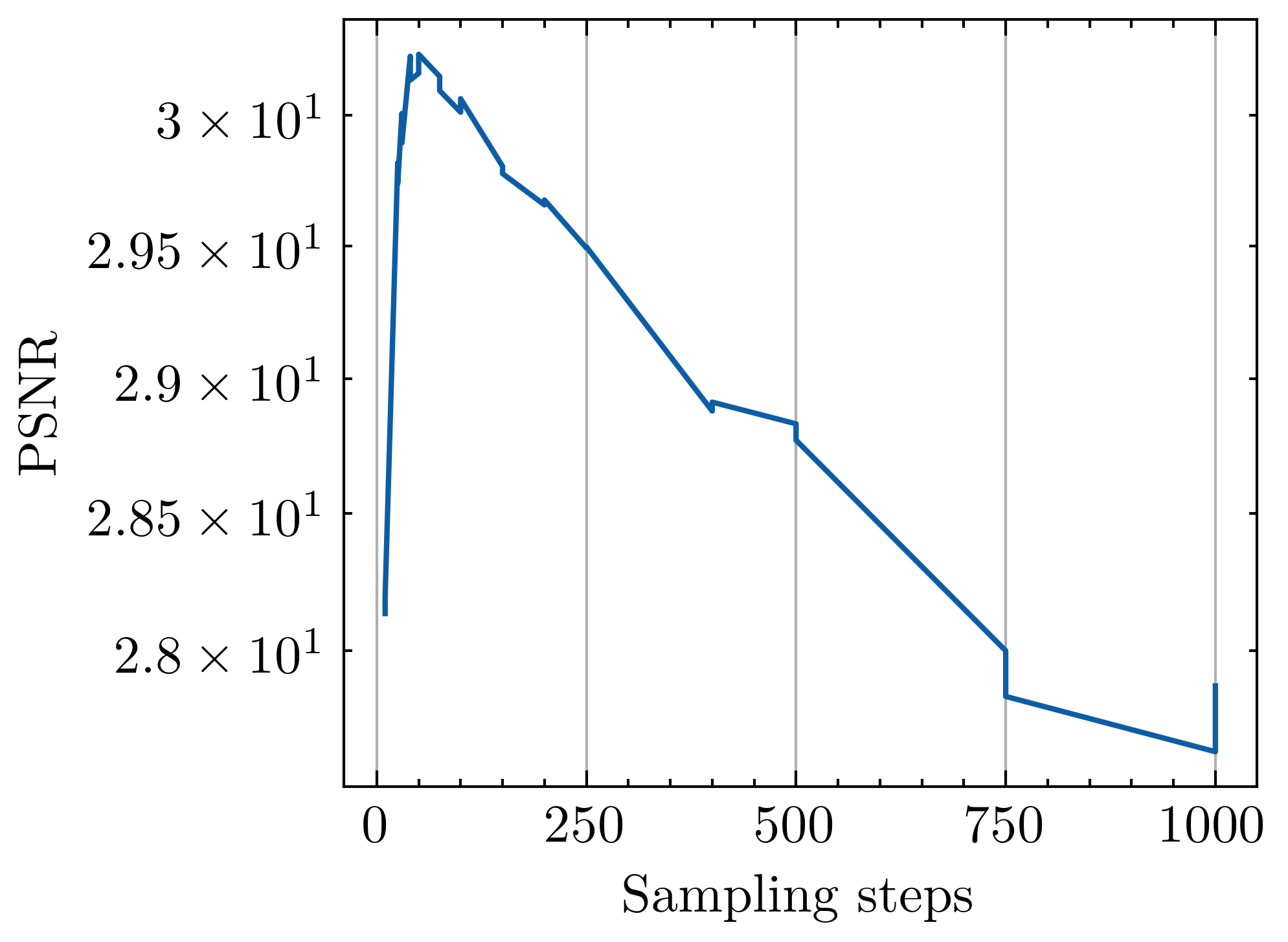}
        \par (c) PSNR
        \label{fig:ffhq_sr_psnr}
    \end{minipage}
    \hspace{1cm}
    \begin{minipage}{0.37\textwidth}
        \centering
        \includegraphics[width=\linewidth]{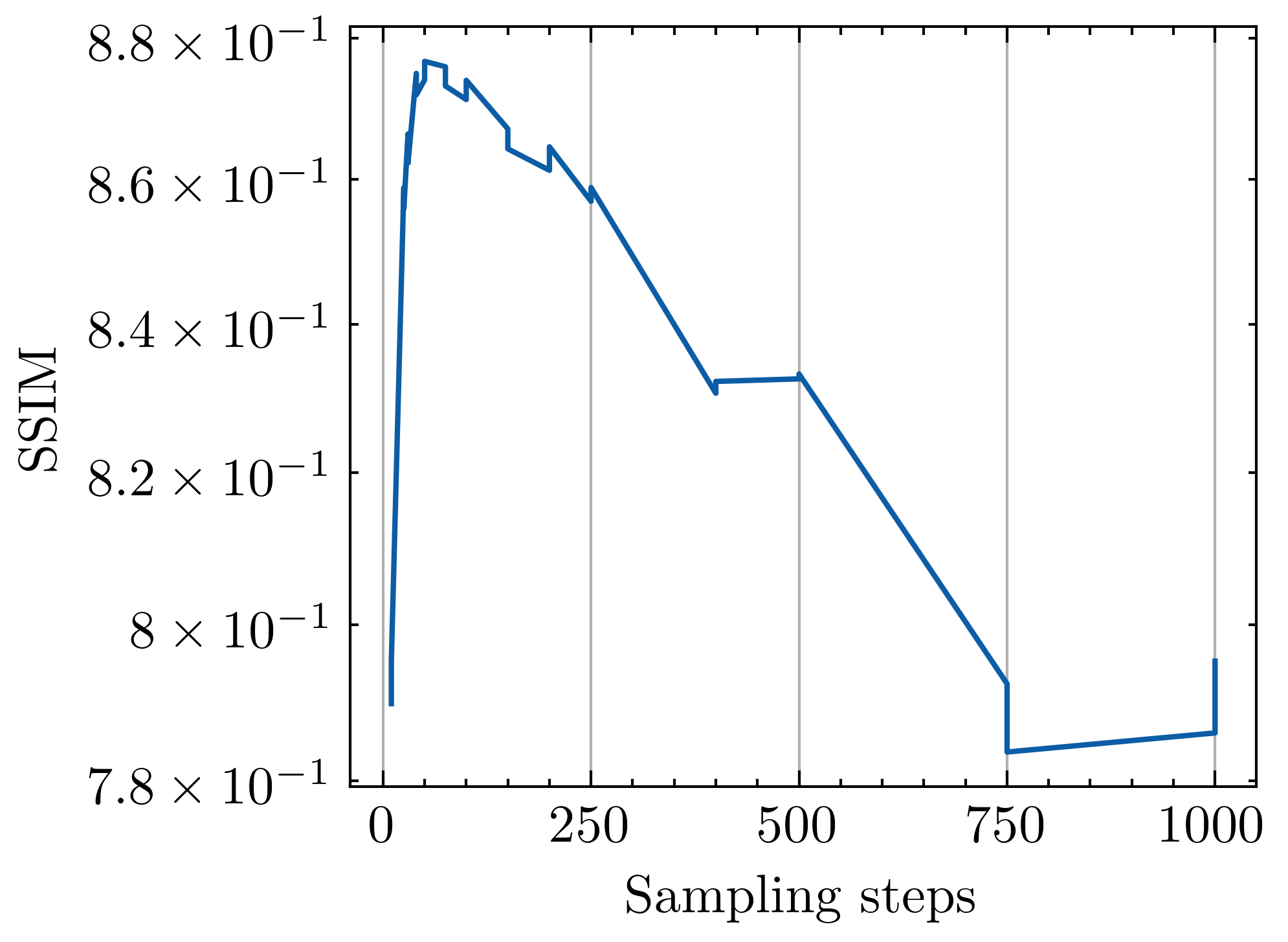}
        \par (d) SSIM 
        \label{fig:ffhq_sr_ssim}
    \end{minipage}

    \caption{Relationship between the evaluation metrics and number of sampling steps on super-resolution task on FFHQ 256 $\times$ 256.}
    \label{fig:ffhq_sr_steps}
\end{figure*}

\begin{figure*}
    \centering

    \begin{minipage}{0.37\textwidth}
        \centering
        \includegraphics[width=\linewidth]{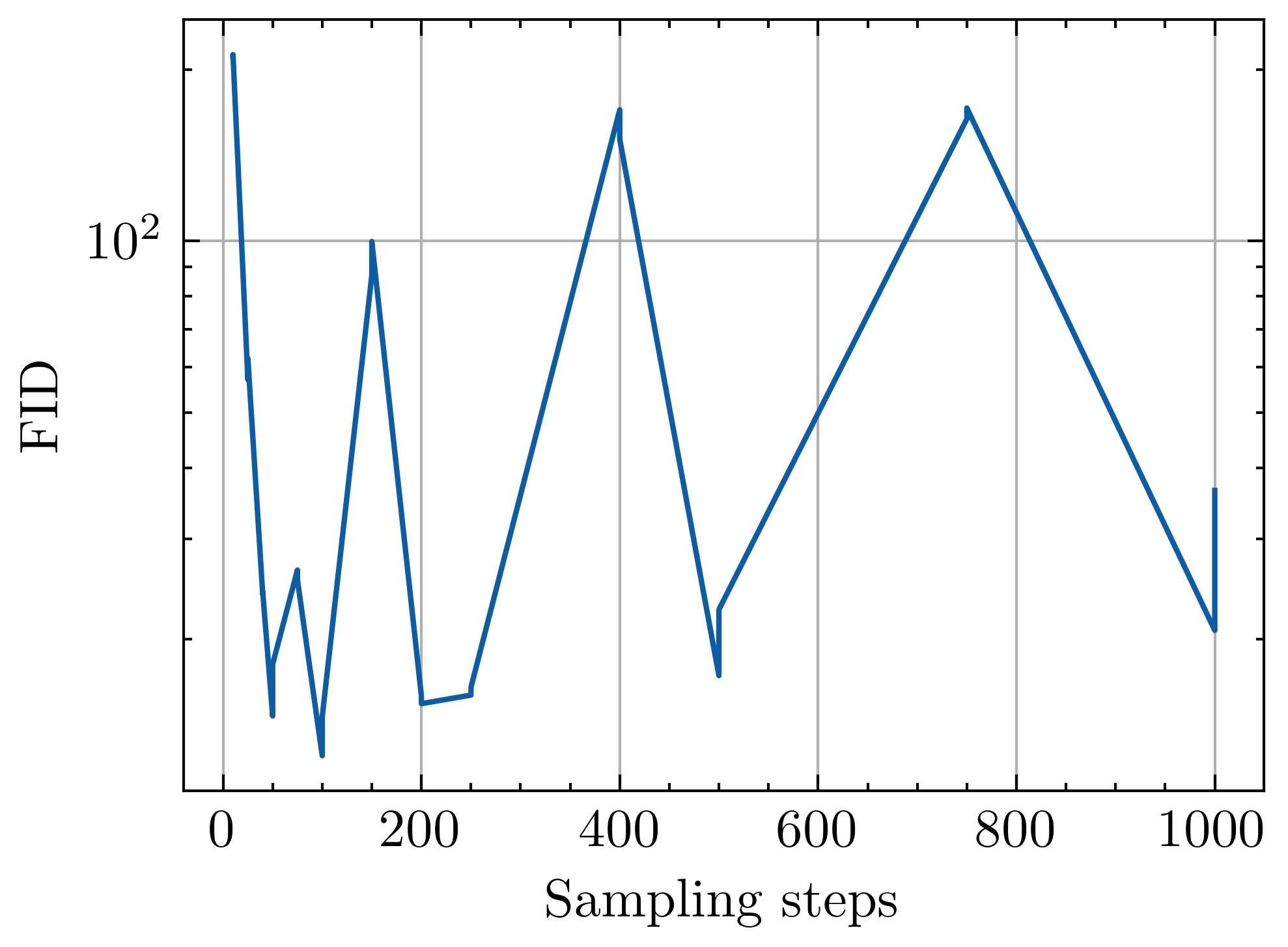}
        \par (a) FID 
        \label{fig:inpaint_fid}
    \end{minipage}
    \hspace{1cm}
    \begin{minipage}{0.37\textwidth}
        \centering
        \includegraphics[width=\linewidth]{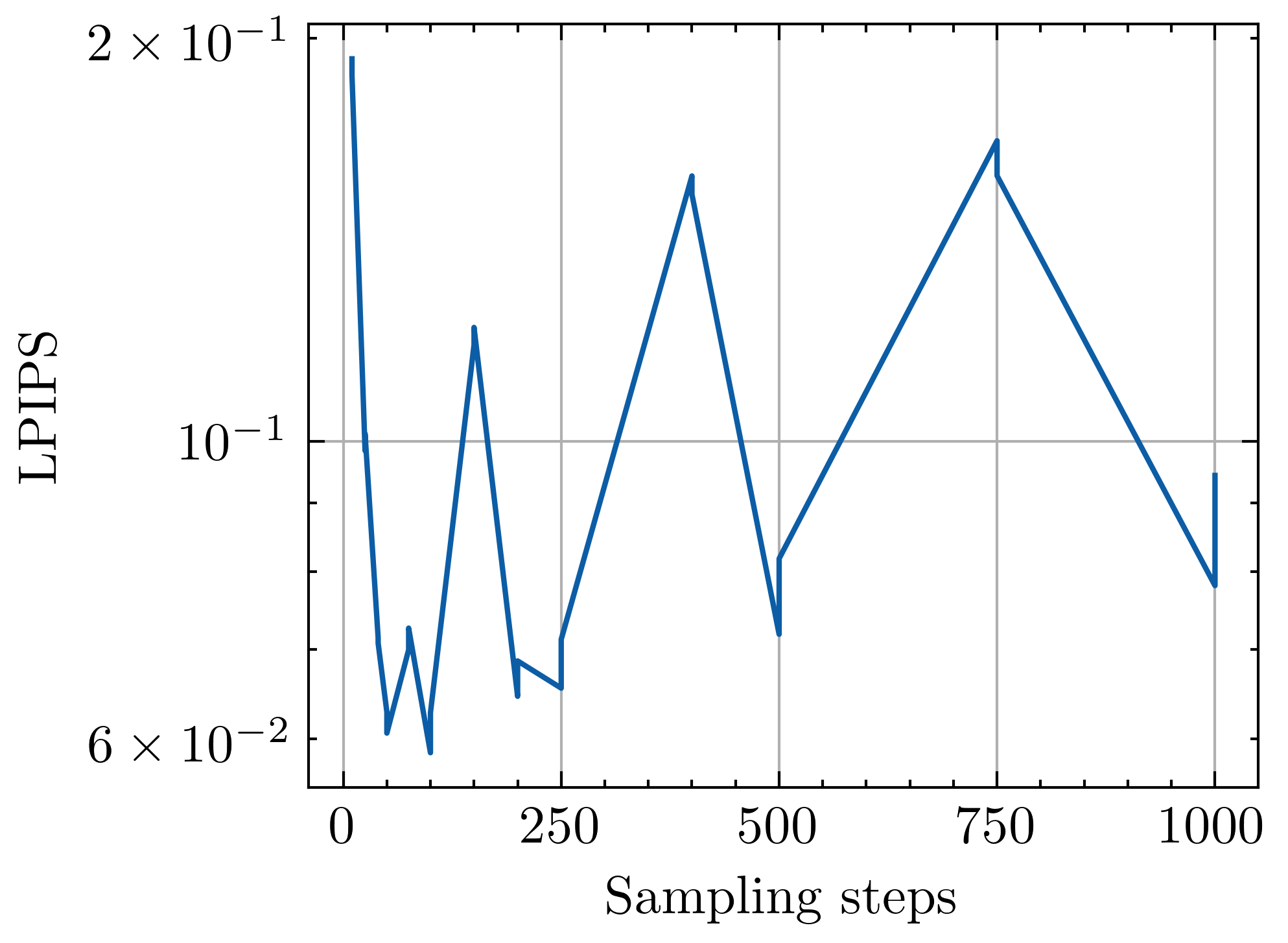}
        \par (b) LPIPS
        \label{fig:inpaint_lpips}
    \end{minipage}

    \begin{minipage}{0.37\textwidth}
        \centering
        \includegraphics[width=\linewidth]{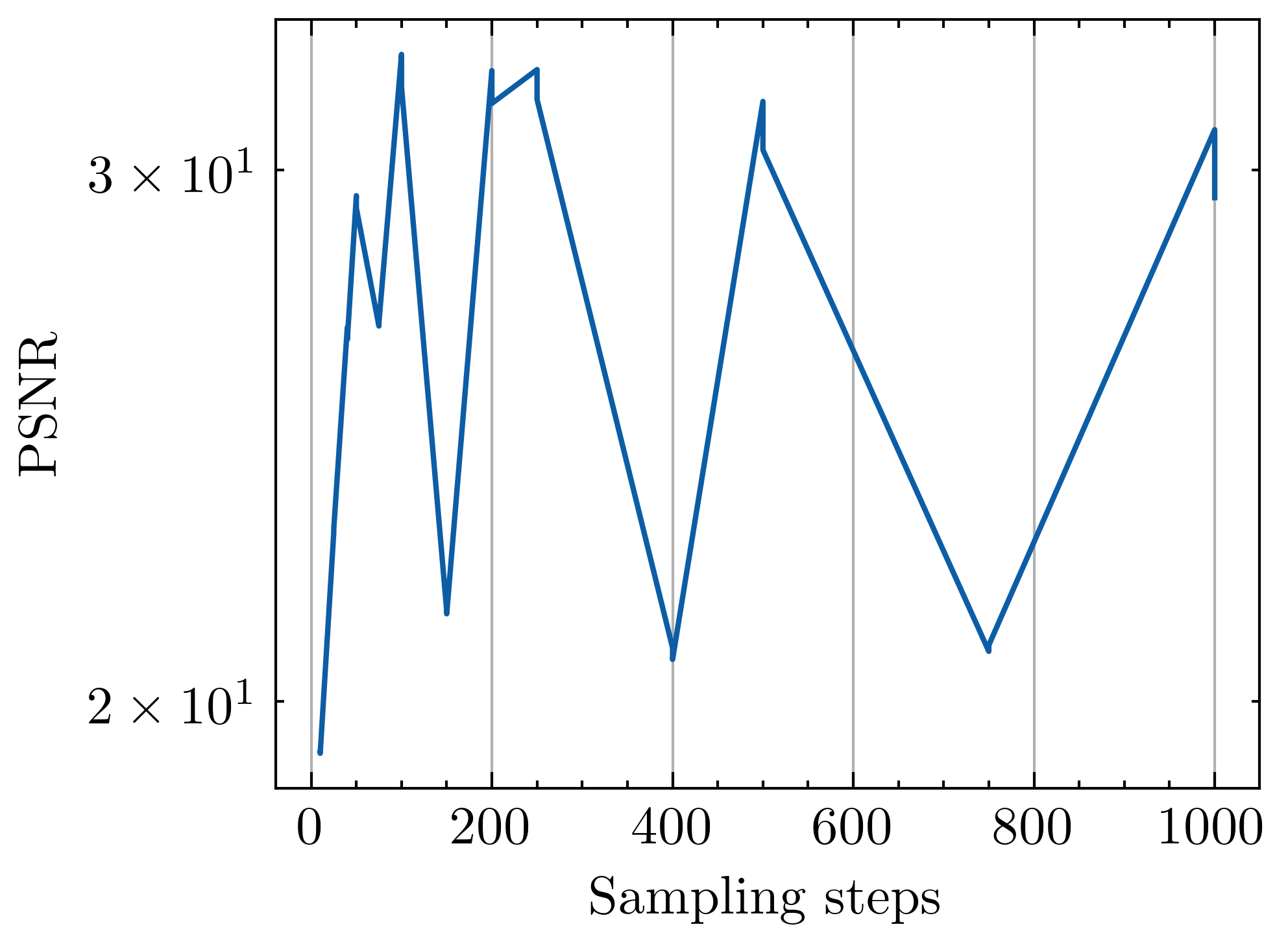}
        \par (c) PSNR
        \label{fig:inpaint_psnr}
    \end{minipage}
    \hspace{1cm}
    \begin{minipage}{0.37\textwidth}
        \centering
        \includegraphics[width=\linewidth]{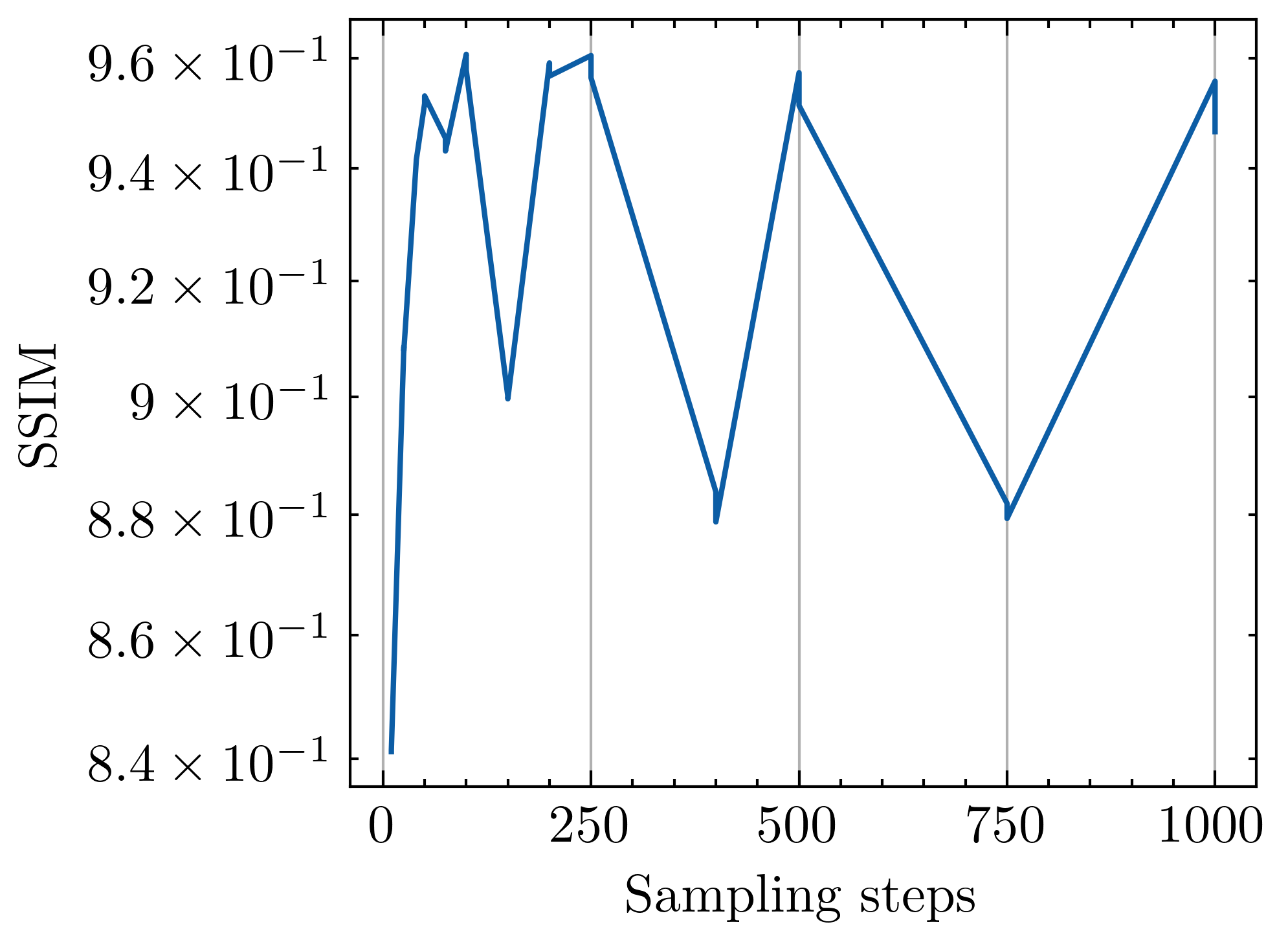}
        \par (d) SSIM 
        \label{fig:inpaint_ssim}
    \end{minipage}

    \caption{Relationship between the evaluation metrics and number of sampling steps on inpainting task on FFHQ $512 \times 512$.}
    \label{fig:inpainting_steps}
\end{figure*}

\begin{figure*}
    \centering
    \begin{minipage}{0.37\textwidth}
        \centering
        \includegraphics[width=\linewidth]{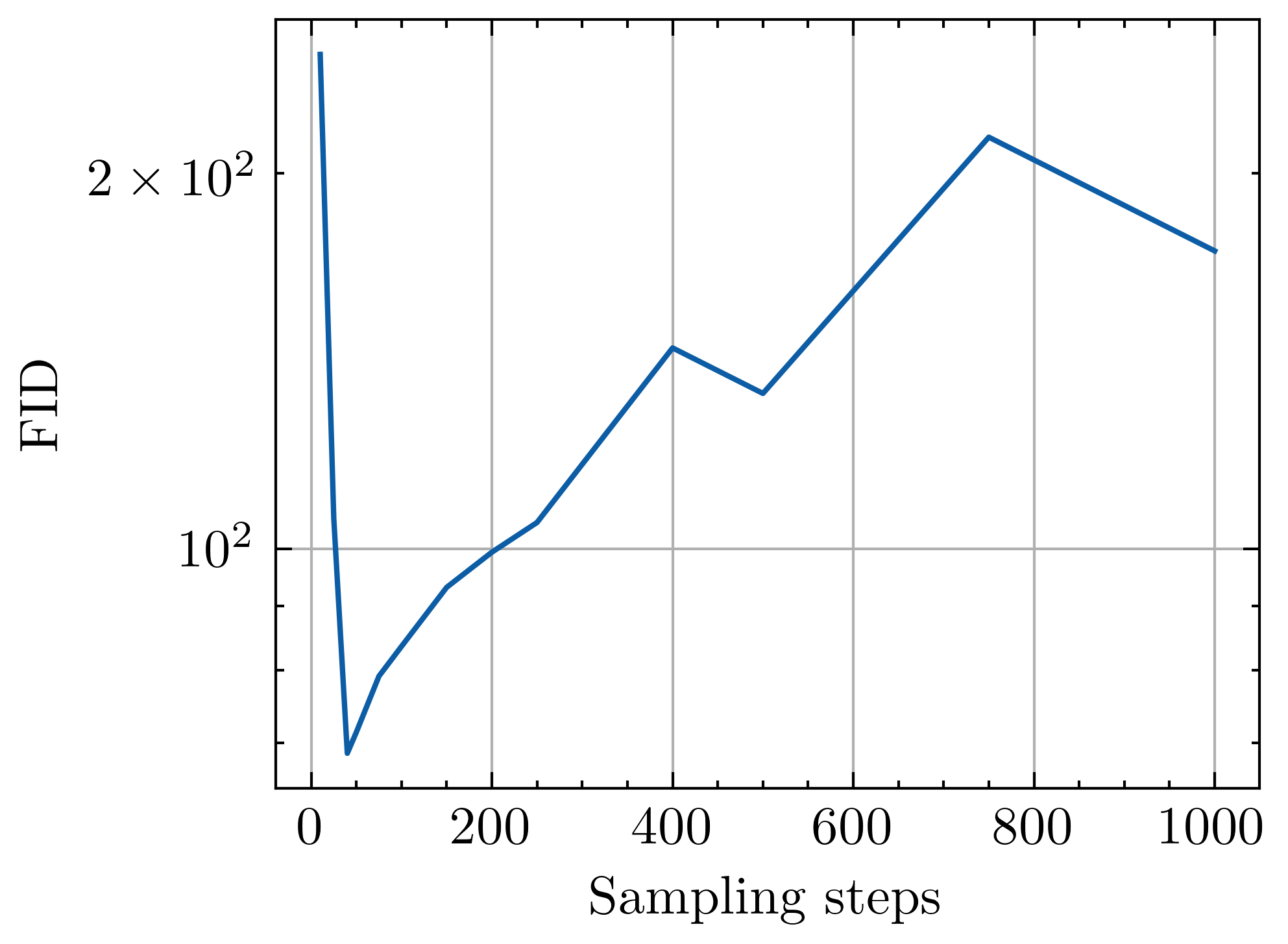}
        \par (a) FID 
        \label{fig:restoration_fid}
    \end{minipage}
    \hspace{1cm}
    \begin{minipage}{0.37\textwidth}
        \centering
        \includegraphics[width=\linewidth]{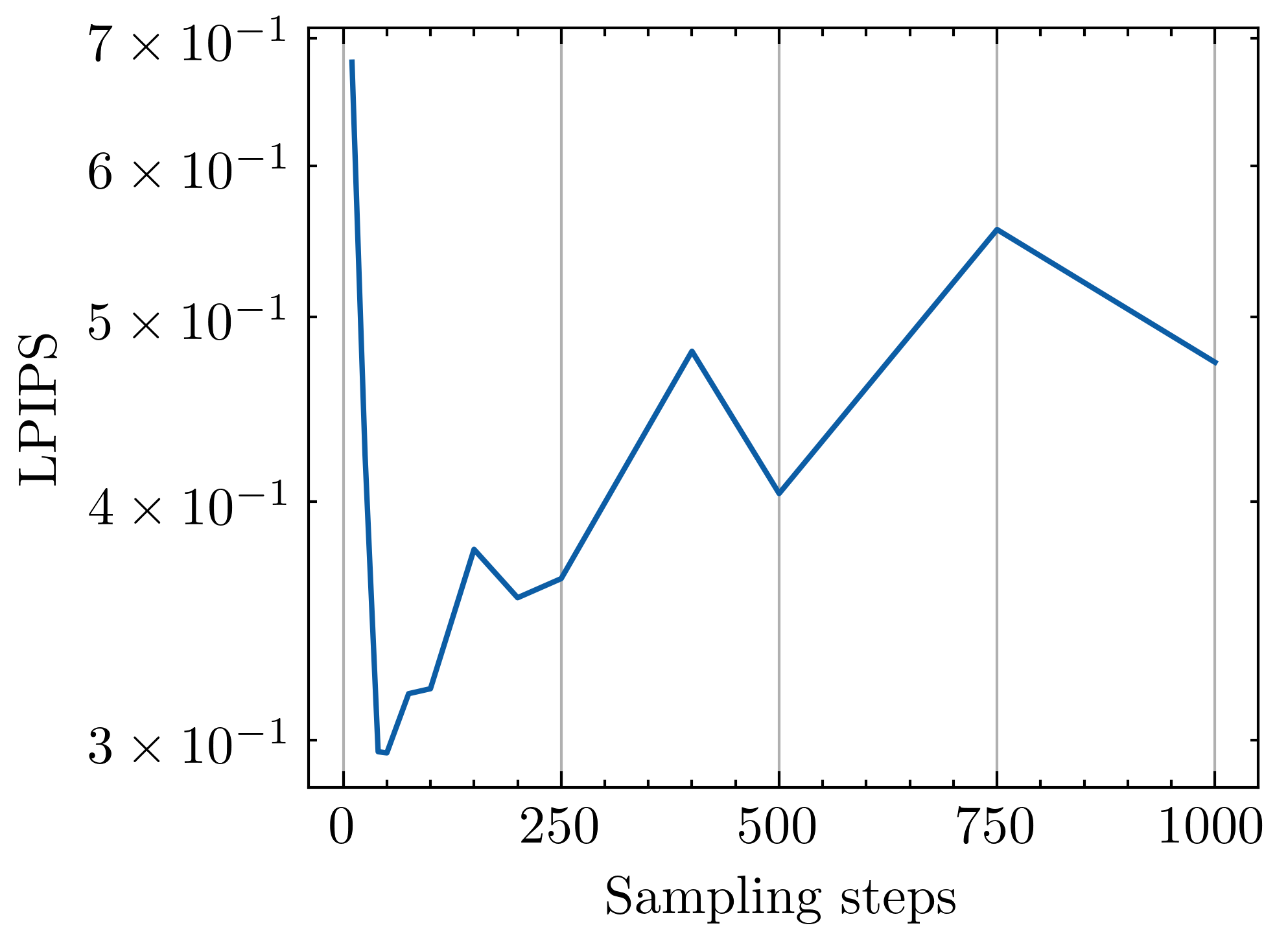}
        \par (b) LPIPS
        \label{fig:restoration_lpips}
    \end{minipage}

    \begin{minipage}{0.37\textwidth}
        \centering
        \includegraphics[width=\linewidth]{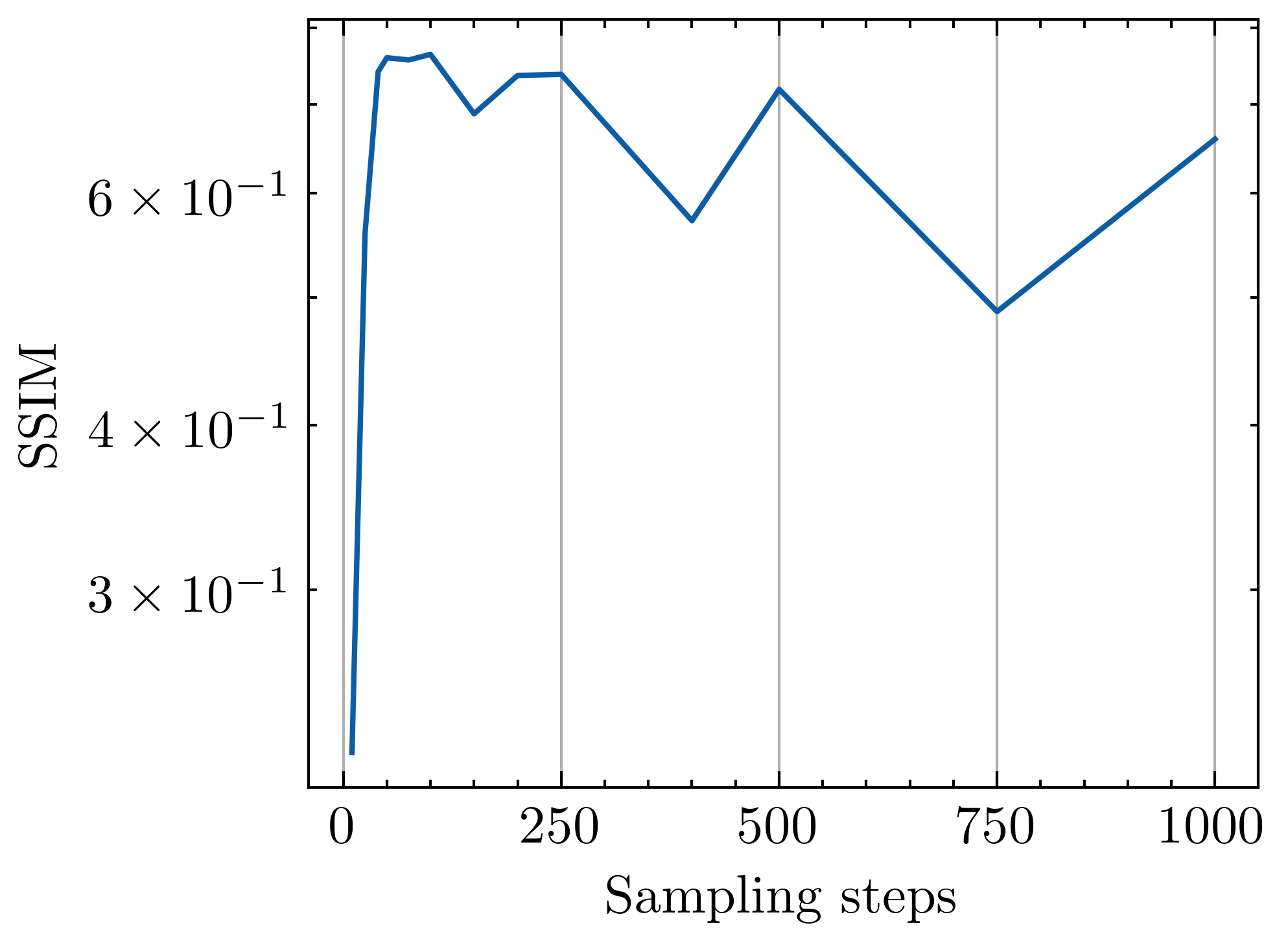}
        \par (c) PSNR
        \label{fig:restoration_psnr}
    \end{minipage}
    \hspace{1cm}
    \begin{minipage}{0.37\textwidth}
        \centering
        \includegraphics[width=\linewidth]{Figures/inpaint_results_paired_ssim_mean_number_of_steps.png}
        \par (d) SSIM 
        \label{fig:restoration_ssim}
    \end{minipage}

    \caption{Relationship between the evaluation metrics and number of sampling steps on blind image restoration task on CelebA 256 $\times$ 256.}
    \label{fig:restoration_steps}
\end{figure*}

The BDPM on image-to-image translation tasks (super-resolution, inpainting and blind image restoration) achieves better results with a smaller number of sampling steps. For every task, we show the relationship between number of sampling steps and evaluation metrics: FID, LPIPS, SSIM and PSNR, in Figures~\ref{fig:ffhq_sr_steps}, \ref{fig:inpainting_steps} and \ref{fig:restoration_steps}. We observe a very similar trend across all tasks: performance improves quickly as sampling steps increase from very few up to a moderate count (tens to low hundreds), reaches its optimum there, and then plateaus or even degrades as the number of steps grows larger. All metrics reported over 1,000 images generated from the validation set.

\subsection{Class-conditional image generation}

\begin{figure*}
    \centering
    \begin{minipage}{0.37\textwidth}
        \centering
        \includegraphics[width=\linewidth]{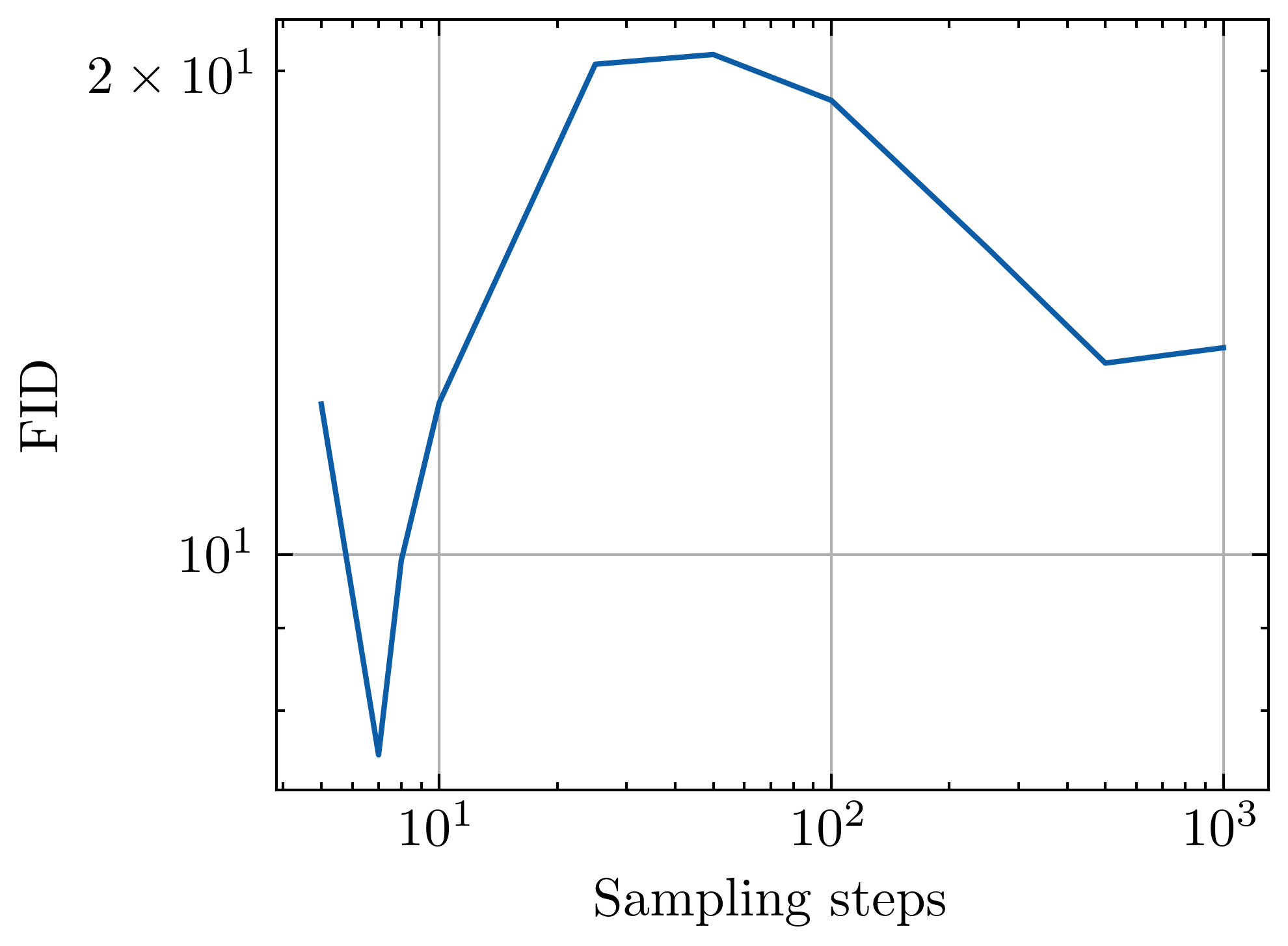}
        \par (a) FID 
        \label{fig:class_cond_fid}
    \end{minipage}
    \hspace{1cm}
    \begin{minipage}{0.37\textwidth}
        \centering
        \includegraphics[width=\linewidth]{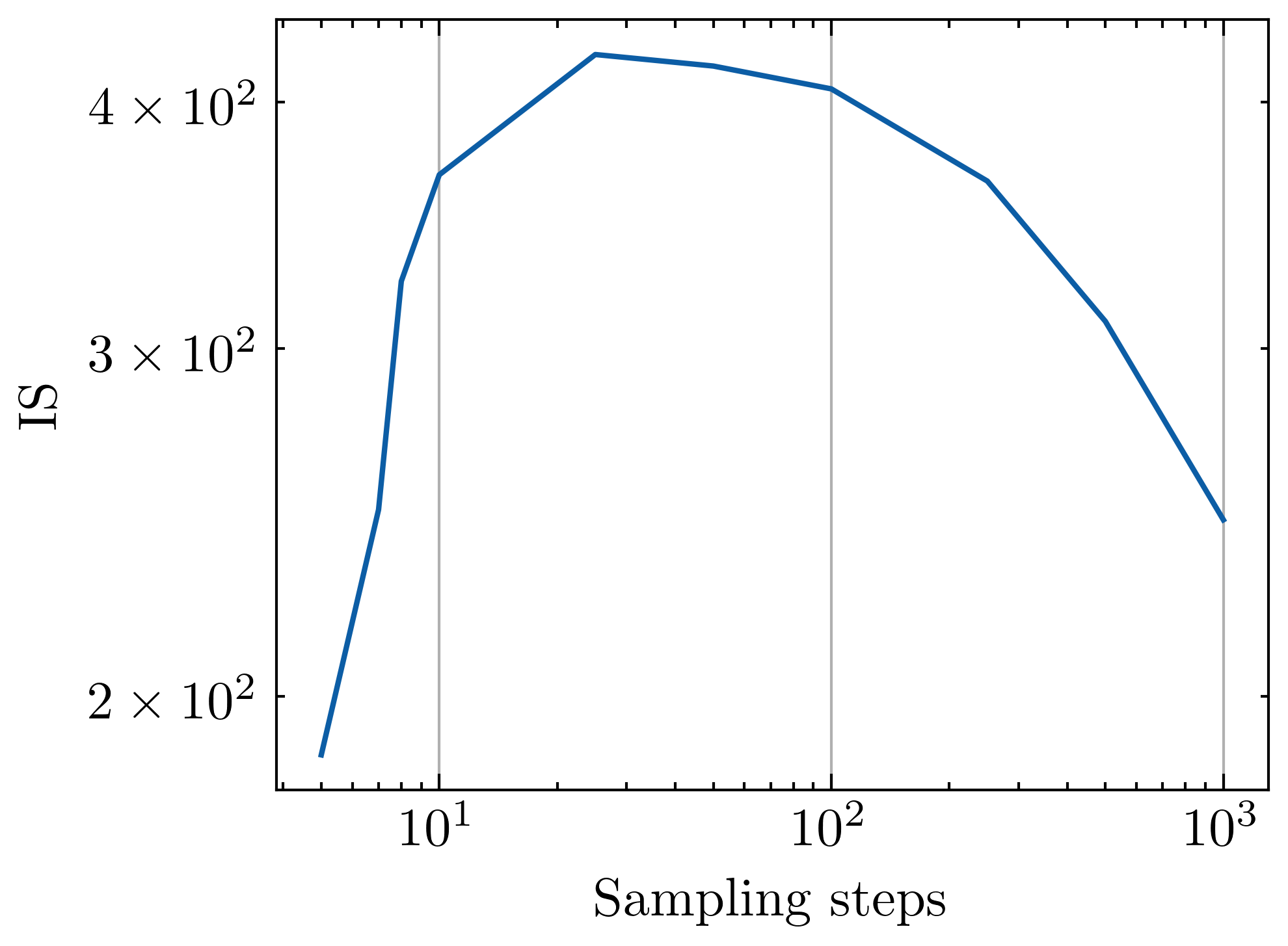}
        \par (b) IS
        \label{fig:class_cond_is}
    \end{minipage}

    \begin{minipage}{0.37\textwidth}
        \centering
        \includegraphics[width=\linewidth]{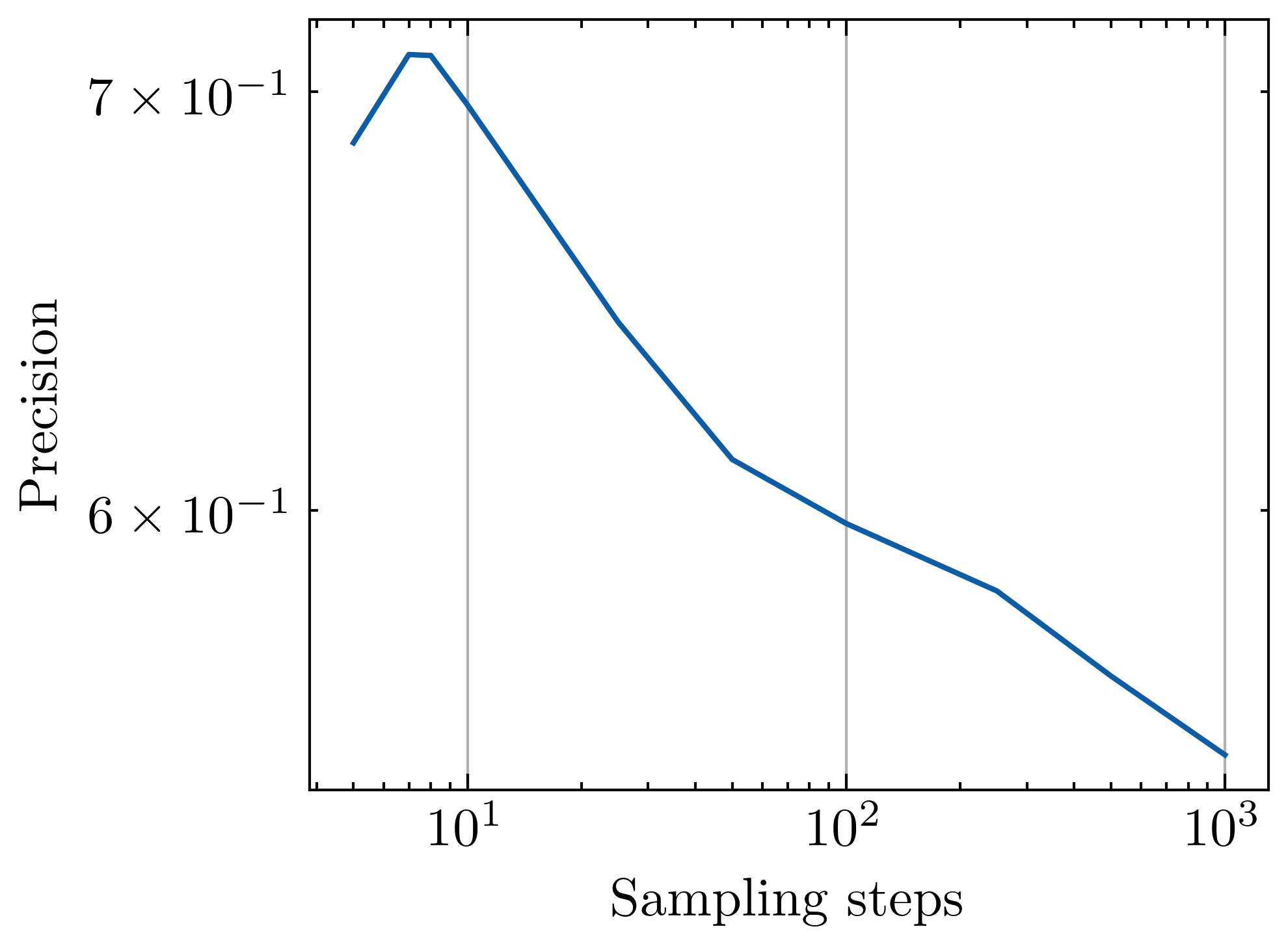}
        \par (c) Precision
        \label{fig:class_cond_precision}
    \end{minipage}
    \hspace{1cm}
    \begin{minipage}{0.37\textwidth}
        \centering
        \includegraphics[width=\linewidth]{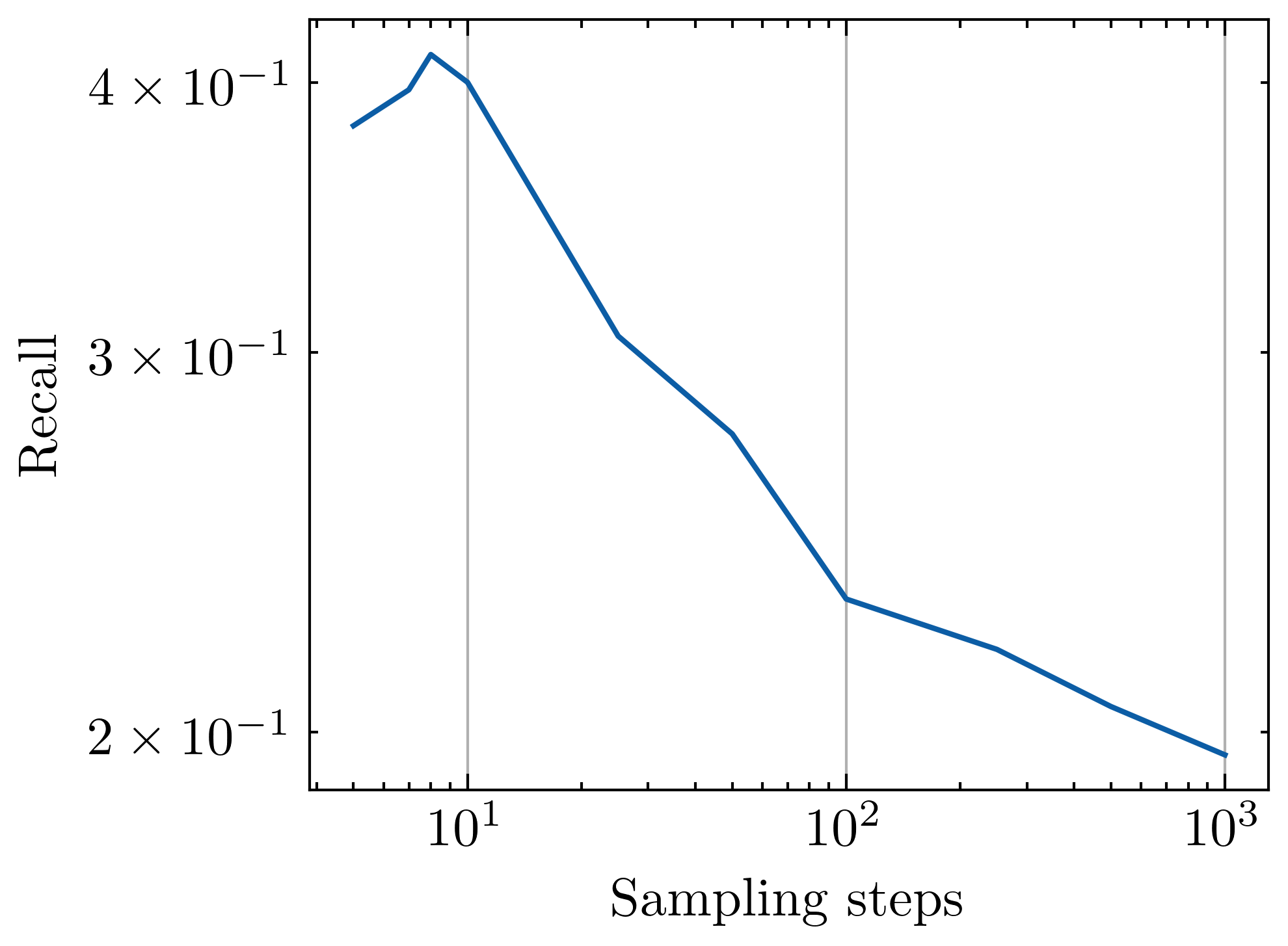}
        \par (d) Recall
        \label{fig:class_cond_recall}
    \end{minipage}

    \caption{Relationship between the evaluation metrics and number of sampling steps on class-conditional image generation on ImageNet-1k 256 $\times$ 256.}
    \label{fig:class_cond_steps}
\end{figure*}

The BDPM (DiT-S, 32.9M parameters) on class-conditional image generation on ImageNet-1k $256\times256$ achieves the best performance at a moderate number of sampling steps. Figure~\ref{fig:class_cond_steps} shows the relationship between sampling steps and evaluation metrics: FID, IS, Precision and Recall. We observe a consistent pattern: as the number of steps increases from very few to tens, all metrics improve rapidly, reach an optimum in the low-to-mid tens of steps, and then gradually decline or plateau as sampling steps grow larger. All metrics are reported on 50,000 images.

\section{Results}

\begin{figure*}
    \centering
    \includegraphics[width=0.65\linewidth]{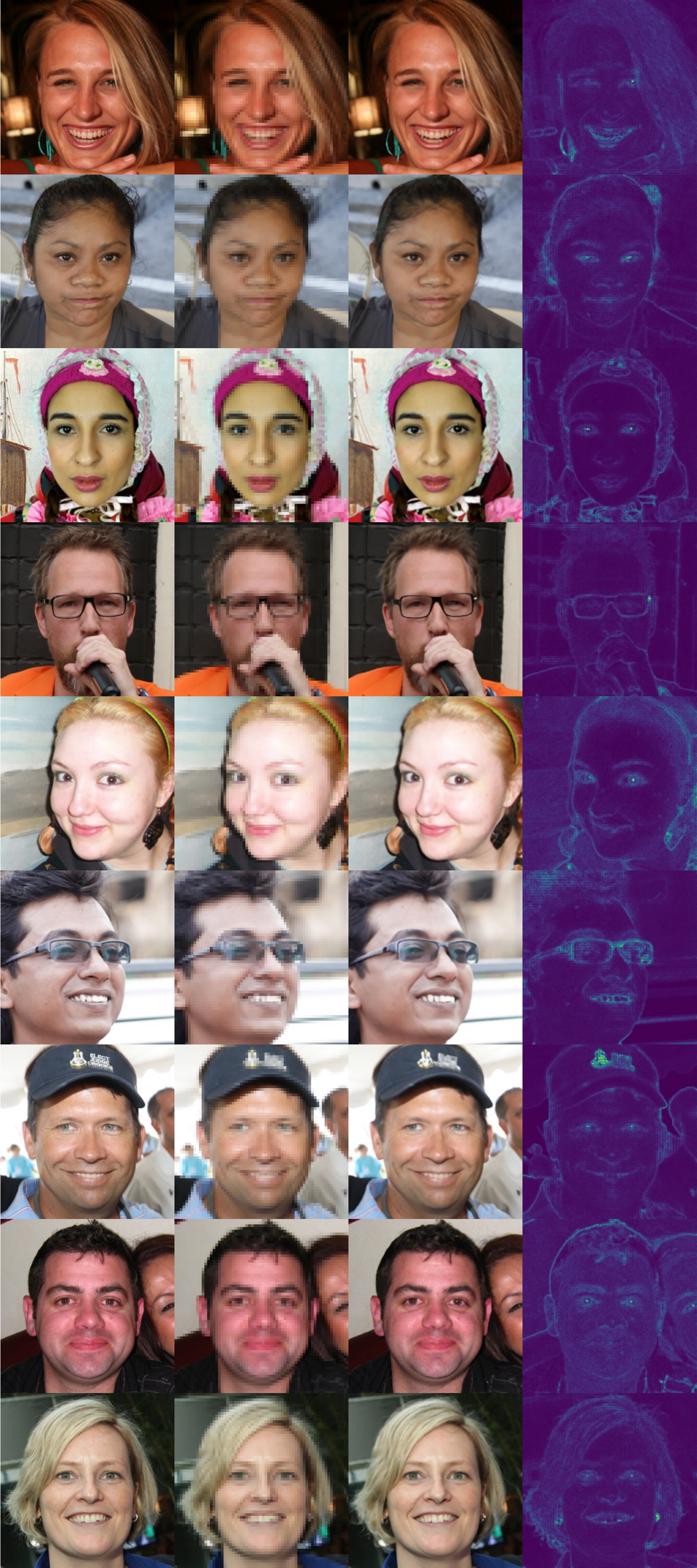}
    \caption{FFHQ super-resolution 256 $\times$ 256. First column: ground truth image, second column: low resolution image, third column: high-resolution image generated by BDPM, forth column: per-pixel variance over 10 high-resolution generations.}
    \label{fig:ffhq_sr_results}
\end{figure*}

\begin{figure*}
    \centering
    \includegraphics[width=0.65\linewidth]{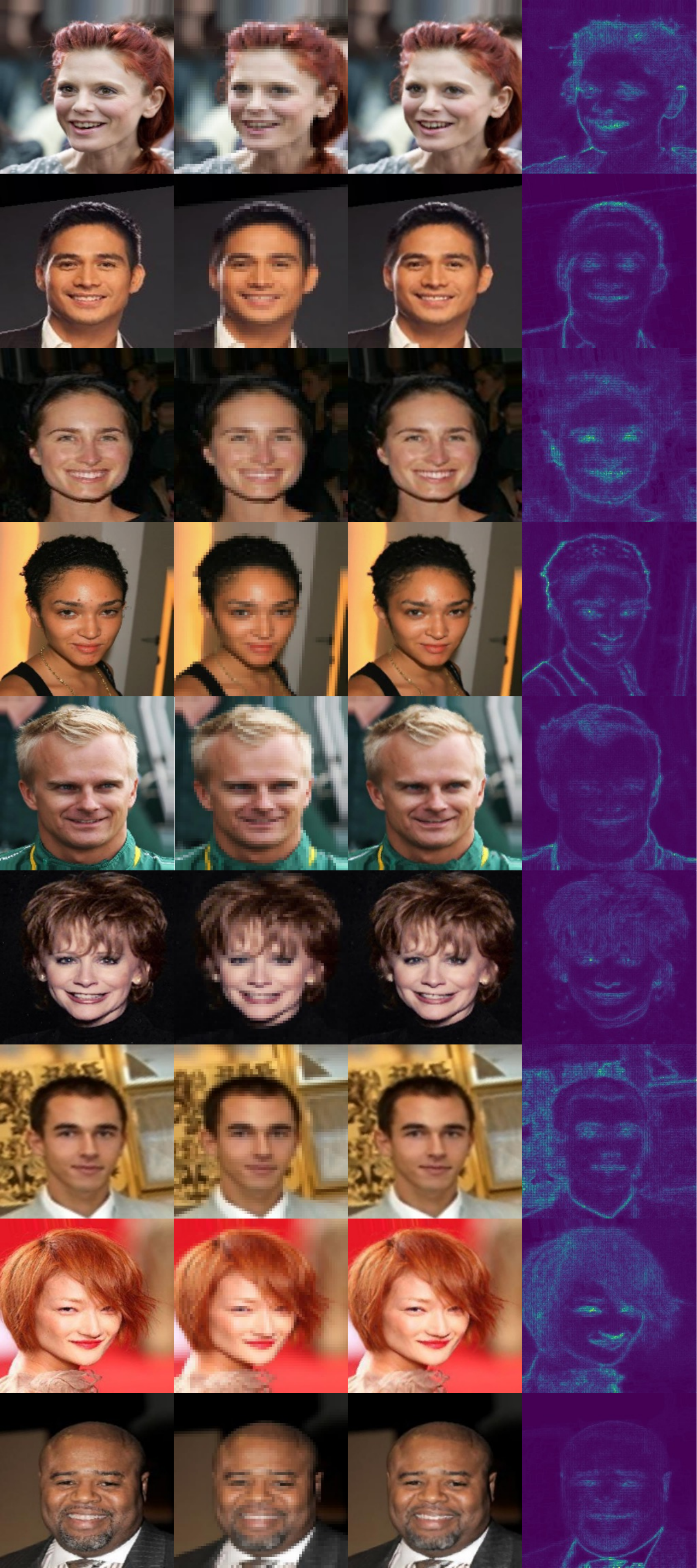}
    \caption{CelebA super-resolution 256 $\times$ 256. First column: ground truth image, second column: low resolution image, third column: high-resolution image generated by BDPM, forth column: per-pixel variance over 10 high-resolution generations.}
    \label{fig:celeba_sr_results}
\end{figure*}

Figures~\ref{fig:ffhq_sr_results},\ref{fig:celeba_sr_results} show super-resolution samples generated using BDPM for datasets FFHQ and CelebA respectively. Ground truth images are shown in the first column, low resolution images are shown in the second columns, high-resolution image generated by BDPM, per-pixel variance over 10 high-resolution generations, per-pixel variances over 10 high-resolution generations are shown in the fourth columns.

\begin{figure*}
    \centering
    \includegraphics[width=0.73\linewidth]{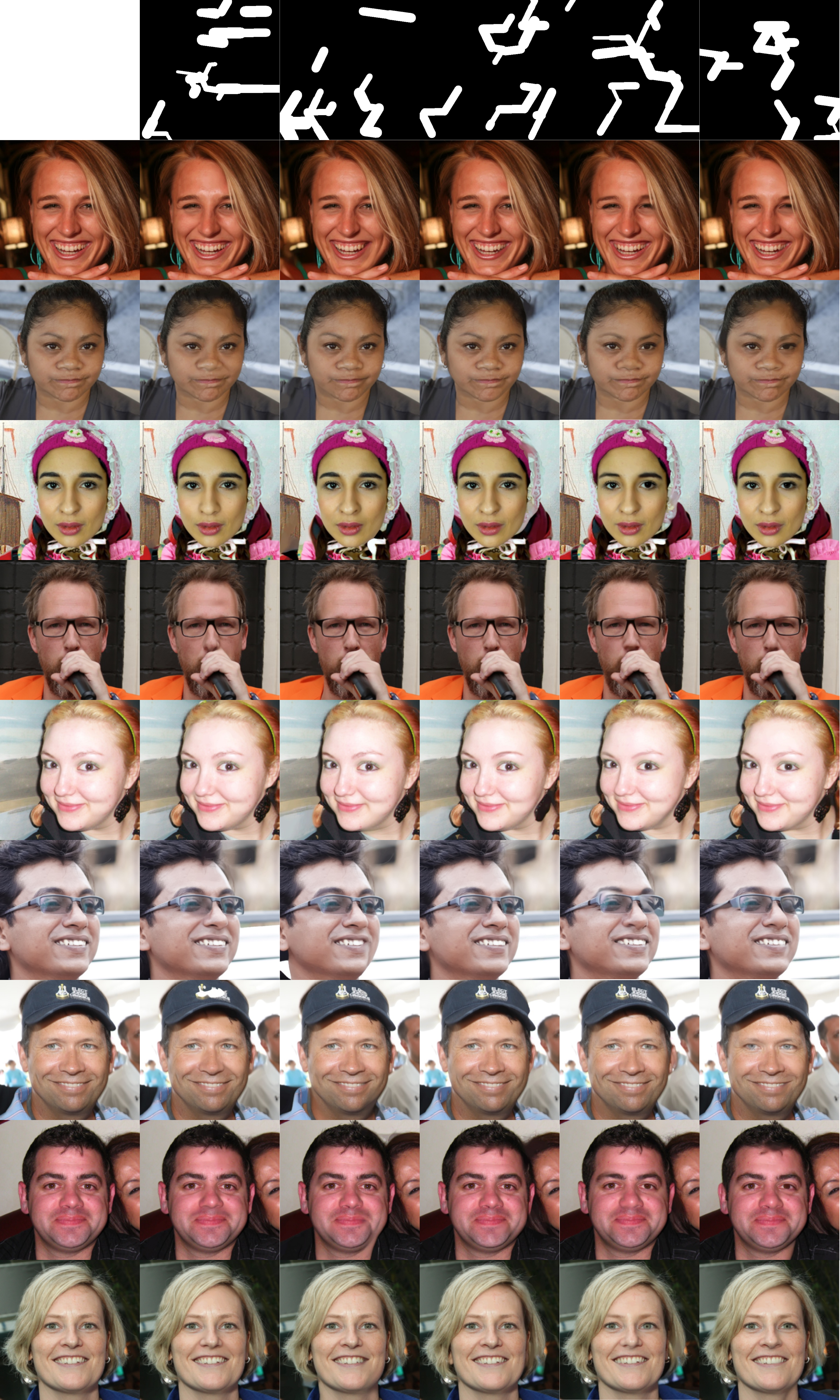}
    \caption{FFHQ inpainting 512 $\times$ 512. First row: inpainting masks, first column: ground truth images, 2-6 columns: inpainting images generated by BDPM.}
    \label{fig:ffhq_inpaint_results}
\end{figure*}

\begin{figure*}
    \centering
    \includegraphics[width=0.73\linewidth]{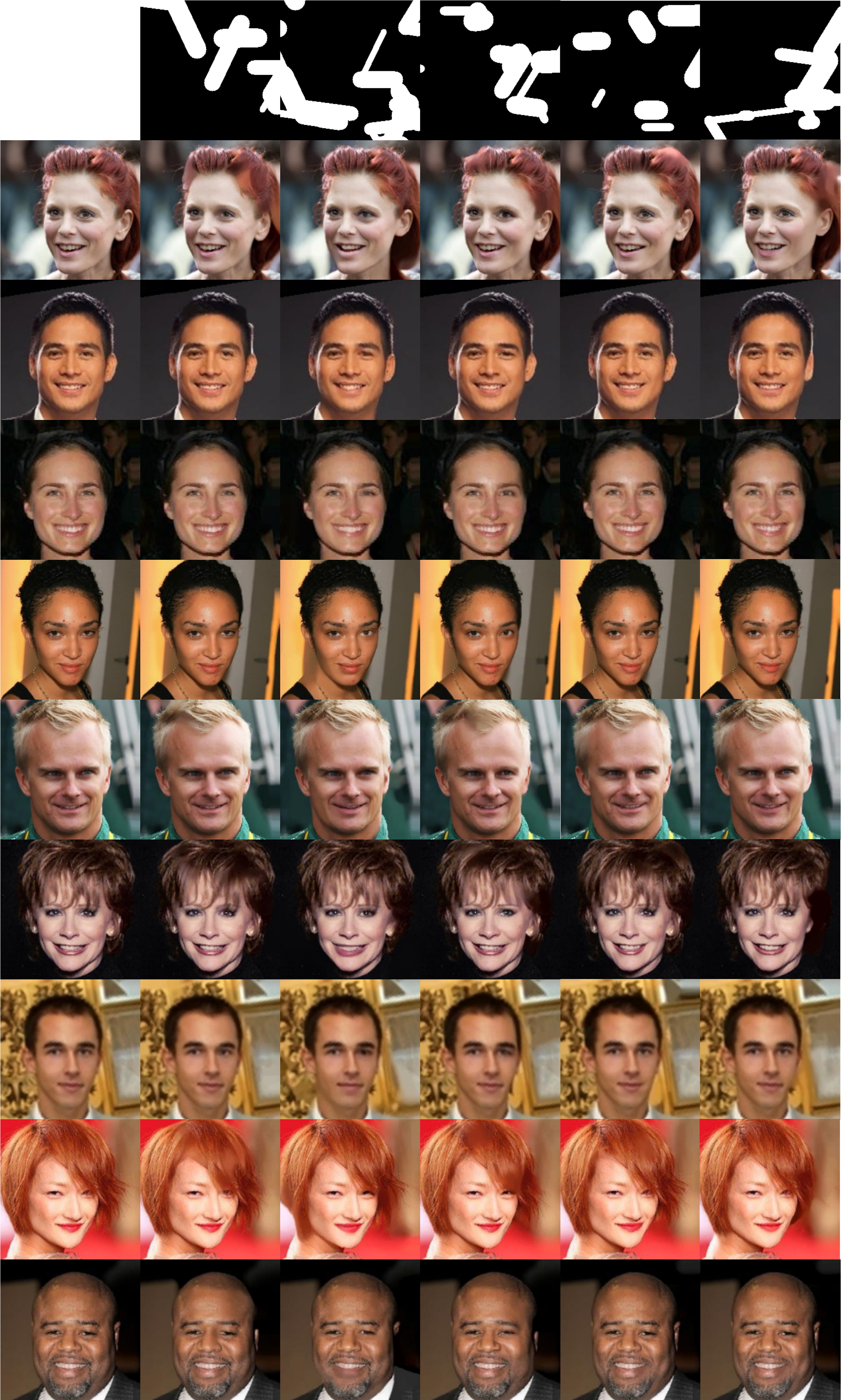}
    \caption{CelebA inpaint 256 $\times$ 256. First row: inpainting masks, first column: ground truth images, 2-6 columns: inpainting images generated by BDPM.}
    \label{fig:celeba_inpaint_results}
\end{figure*}

\begin{figure*}
    \centering
    \includegraphics[width=0.73\linewidth]{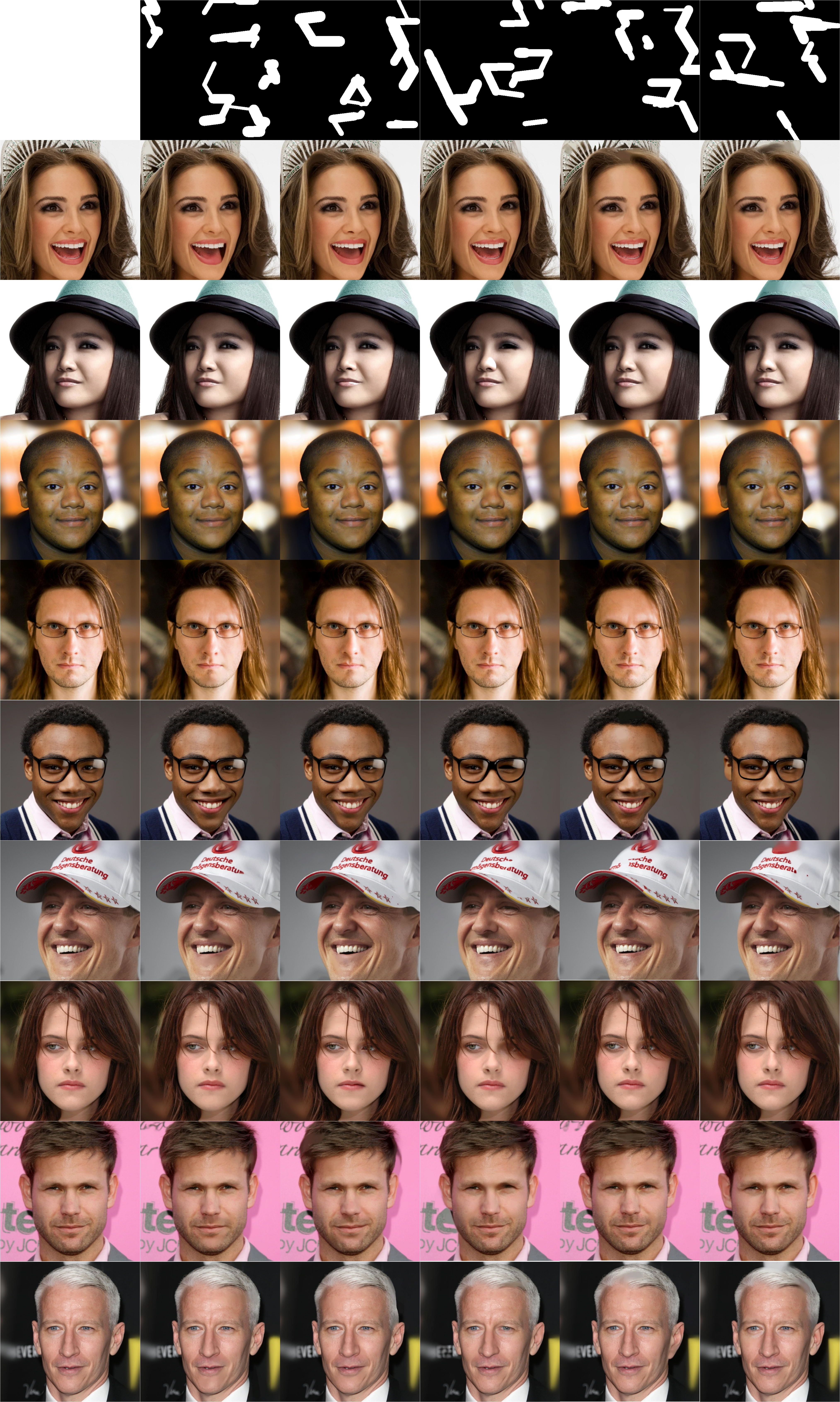}
    \caption{CelebA-HQ inpaint 512 $\times$ 512. First row: inpainting masks, first column: ground truth images, 2-6 columns: inpainting images generated by BDPM.}
    \label{fig:celeba_hq_inpaint_results}
\end{figure*}

Figures~\ref{fig:ffhq_inpaint_results},\ref{fig:celeba_inpaint_results},\ref{fig:celeba_hq_inpaint_results} show inpainting samples generated using BDPM for datasets FFHQ, CelebA and CelebA-HQ respectively. Masks are shown in the first row, ground truth images are shown in the first column, inpainting samples for each masks are shown in the columns 2 - 6.

\begin{figure*}
    \centering
    \includegraphics[width=0.68\linewidth]{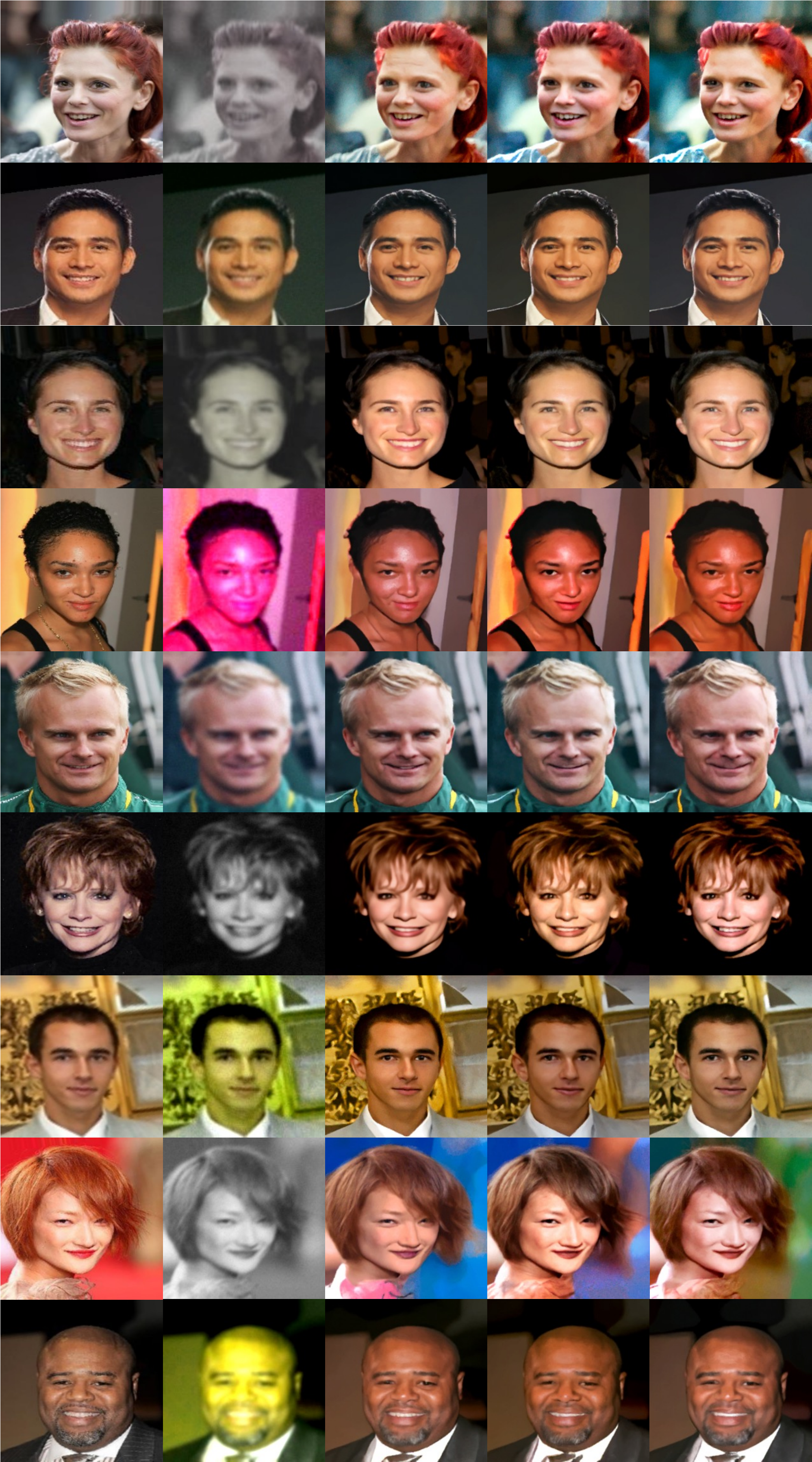}
    \caption{CelebA blind image restoration 256 $\times$ 256. First column: ground truth images, second column: distorted images, 3-5 columns: images restored by BDPM.}
    \label{fig:celeba_restoration_results}
\end{figure*}

Figure~\ref{fig:celeba_restoration_results} shows blind image restortion results. Ground truth images are shown in the first column, perturbed images are shown in the second column, restored images are shown in columns 3 - 5.

\begin{figure*}
    \centering
    \includegraphics[width=0.75\linewidth]{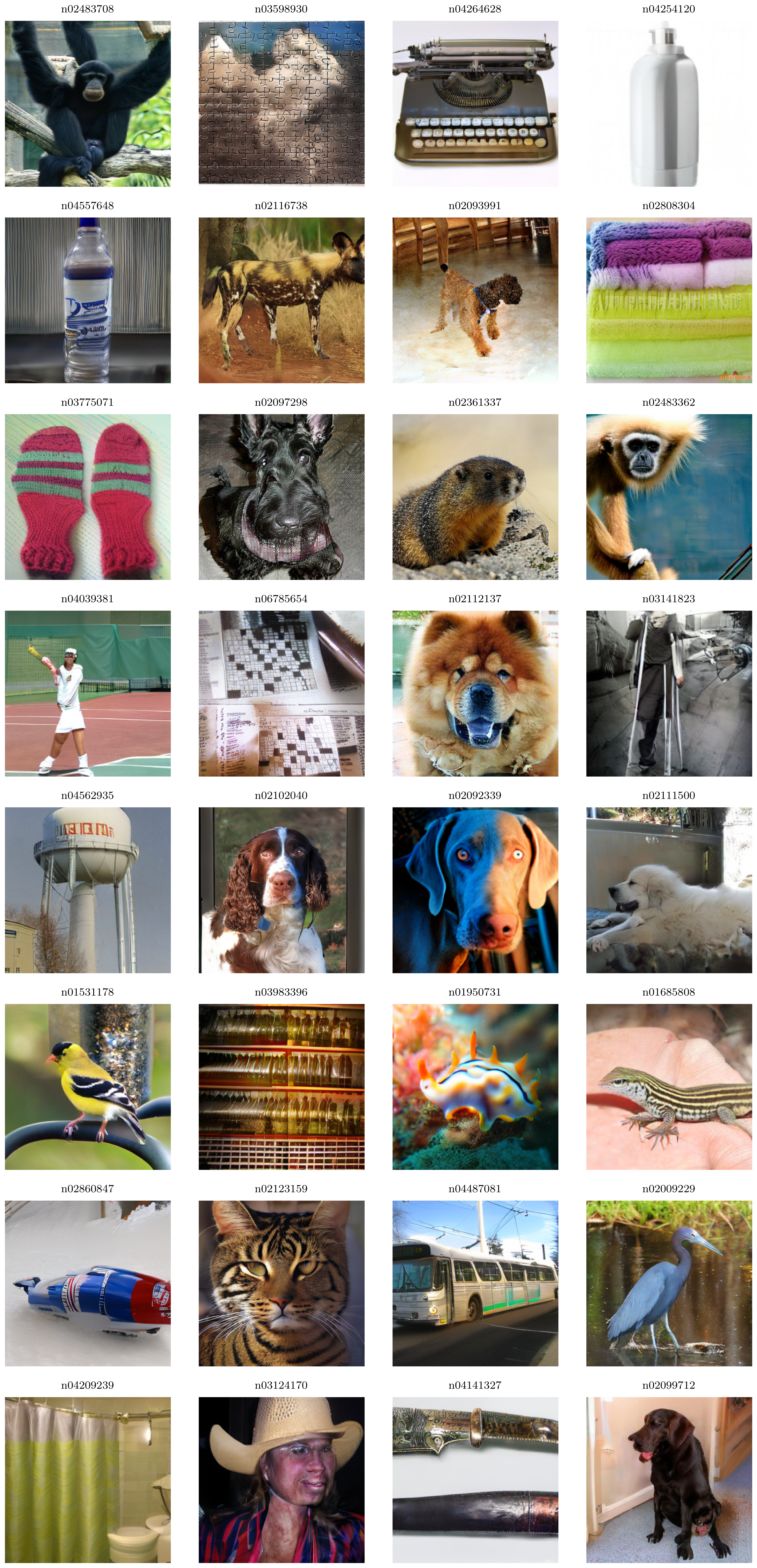}
    \caption{Non-cherry picked class-conditionaly generated ImageNet-1k images by BDPM DiT-S model (32.9M parameters). 7 sampling steps. 11.25 guidance scale.}
    \label{fig:imagenet_s}
\end{figure*}

\begin{figure*}
    \centering
    \includegraphics[width=0.75\linewidth]{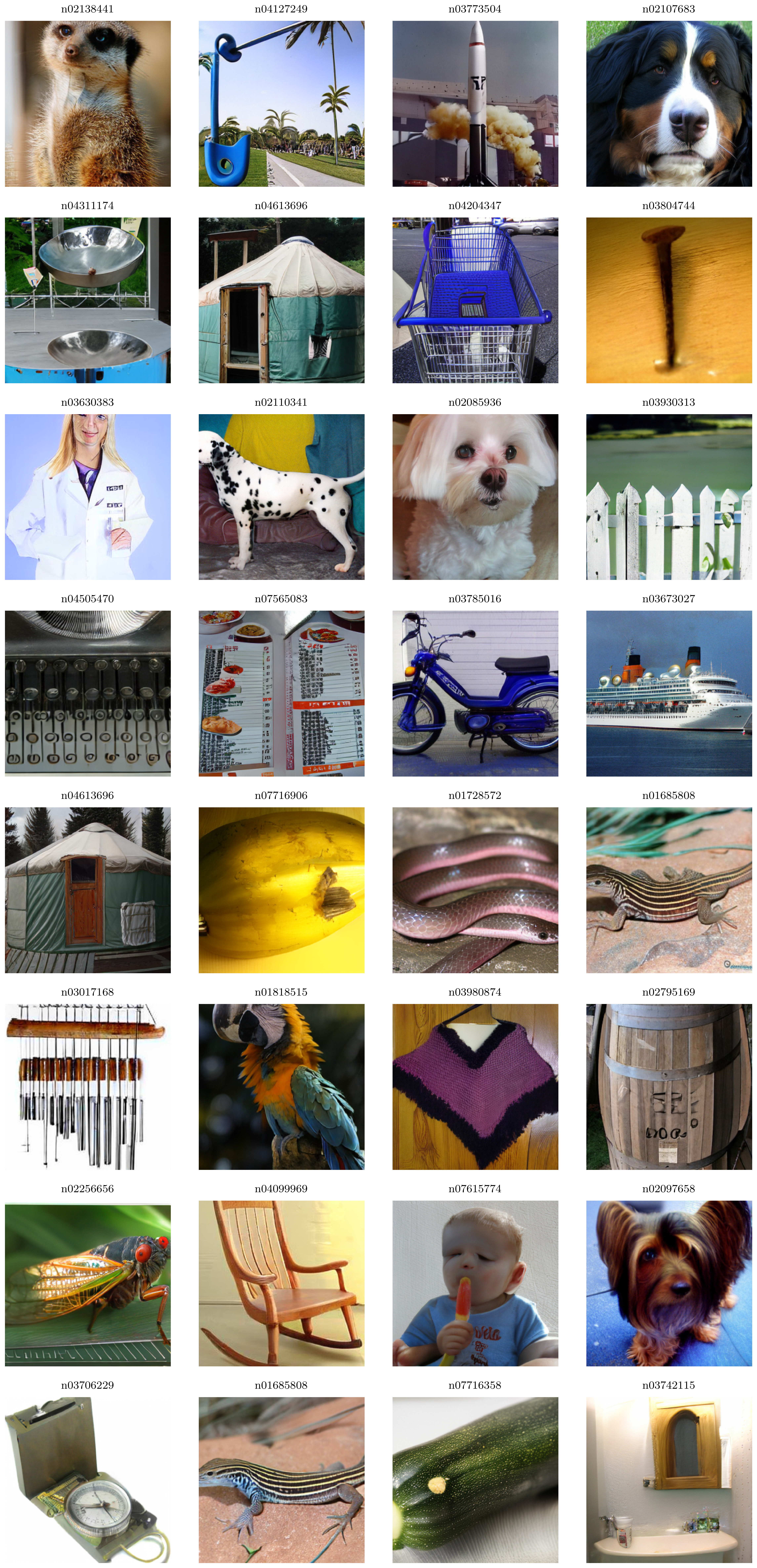}
    \caption{Non-cherry picked class-conditionaly generated ImageNet-1k images by BDPM DiT-B model (130M parameters). 7 sampling steps. 8.75 guidance scale.}
    \label{fig:imagenet_b}
\end{figure*}

Figures~\ref{fig:imagenet_s},\ref{fig:imagenet_b} show non-cherry picked class-conditionaly generated ImageNet-1k samples. The class label is provided for every image on top of it.

\end{document}